\documentclass[11pt]{article}

\usepackage[final]{acl}
\usepackage{times}
\usepackage{latexsym}

\usepackage[T1]{fontenc}

\usepackage[utf8]{inputenc}

\usepackage{microtype}

\usepackage{multirow}
\usepackage{booktabs}
\usepackage{graphicx}
\usepackage{tikz}
\usetikzlibrary{shapes.geometric, positioning, calc}
\usepackage{amsmath}
\usepackage{amsthm}
\usepackage{booktabs}
\usepackage{algorithm}
\usepackage{algorithmic}
\usepackage[switch]{lineno}

\usepackage{arydshln}
\usepackage{xcolor}
\usepackage{colortbl}

\usepackage{booktabs} 
\usepackage{multirow}
\usepackage{subcaption}

\usepackage{enumitem}
\usepackage{graphicx}
\usepackage{xcolor}    
\usepackage{listings}
\usepackage[most]{tcolorbox} 
\tcbset{
  aibox/.style={
    width=474.18663pt,
    top=10pt,
    colback=white,
    colframe=black,
    colbacktitle=black,
    enhanced,
    center,
    attach boxed title to top left={yshift=-0.1in,xshift=0.15in},
    boxed title style={boxrule=0pt,colframe=white,},
  }
}
\newtcolorbox{promptRed}[2][]{
  aibox,
  colback=violet!3,
  colframe=violet!10!black,
  colbacktitle=violet!10!black,
  title=#2,
  breakable,  
  listing engine=listings,
  #1
}

\newtcolorbox{promptBlue}[2][]{
  aibox,
  colback=blue!3,
  colframe=blue!10!black,
  colbacktitle=blue!10!black,
  title=#2,
  #1
}

\NewDocumentCommand{\hongru}
{ mO{} }{\textcolor{red}{\textsuperscript{\textit{Hongru}}\textsf{\textbf{\small[#1]}}}}

\usepackage{pifont} 
\newcommand{\cmark}{\ding{51}} 
\newcommand{\xmark}{\ding{55}} 

\newcommand{\our}{{\textit{\textbf{KA}\scalebox{0.8}{\textbf{ware}}}}}

\usepackage{multirow}

\title{\textbf{\textit{From Knowing to Acting:}} Benchmarking Self-Awareness Capability of \\ LLM Agents}

\author{\textbf{Yifan Li\textsuperscript{1}\protect\footnotemark[2]},
 \textbf{Shengbin Yue\textsuperscript{2}\protect\footnotemark[2]}, 
  \textbf{Boyu Feng\textsuperscript{1}\protect\footnotemark[2]}, 
 \textbf{Jinhu Qi\textsuperscript{1}},
  \textbf{Bo Ke\textsuperscript{4}}, \\
  \textbf{Zixing Song\textsuperscript{5}},
 \textbf{Hongru Wang\textsuperscript{3}} ,   
 \textbf{Zhongyu Wei\textsuperscript{2}\protect\footnotemark[1]},
 \textbf{Irwin King\textsuperscript{1}\protect\footnotemark[1]}
 \\
 \textsuperscript{1}The Chinese University of Hong Kong
\textsuperscript{2}Fudan University \\
  \textsuperscript{3}University of Edinburgh
 \textsuperscript{4}Tencent 
  \textsuperscript{5}University of Bristol \\
\normalsize\texttt{yfli24@cse.cuhk.edu.hk, sbyue23@m.fudan.edu.cn}
\normalsize\texttt{
}}

\begin{document}
\maketitle
\begin{abstract}
The integration of external tools has transitioned LLM agents from passive responders to autonomous systems. However, current benchmarks prioritize execution success, neglecting self-awareness capability, the ability to discern whether a problem requires necessary external resources or can be solved via internal parametric knowledge. To address this, we introduce {\textit{\textbf{KAP}\scalebox{0.8}{\textbf{RO}}}} (\textit{K}nowing–\textit{A}cting Quadrant \textit{P}\textit{RO}be), a framework that evaluates cognitive-behavioral alignment by decoupling an agent's metacognitive judgment (\textit{Knowing}) from its spontaneous execution (\textit{Acting}). We further construct {\textit{\textbf{KA}\scalebox{0.8}{\textbf{ware}}}}, a dataset rigorously partitioning tasks into external, internal, and hybrid subspaces to systematically probe these epistemic boundaries. Extensive experiments across diverse agent architectures show that self-awareness capability is strongly correlated with task success but degrades sharply in internal-capability settings. Moreover, open-source and instruction-following models exhibit stronger tool overuse due to shallow pattern matching, while proprietary and reasoning-oriented models demonstrate more reliable cognitive gating. Benchmark and codes are available at \url{https://github.com/AI-Santiago/KAware}.
\end{abstract}

\renewcommand{\thefootnote}{\fnsymbol{footnote}}
\footnotetext[2]{These authors contributed equally to this work.}
\footnotetext[1]{Corresponding author}
\renewcommand{\thefootnote}{\arabic{footnote}}

\section{Introduction}
Recent progress in large language models (LLMs) has accelerated their transition from passive dialogue systems toward autonomous agentic assistants~\cite{yao2023react, wu2025webdancer,he2025vitabench}. A hallmark of this shift is the coexistence of two complementary capability sources: \textbf{internal capabilities} (\textit{e.g.}, reasoning, planning, self-reflection) and \textbf{external tools} (\textit{e.g.}, search engines, and databases). Rather than relying solely on parametric knowledge, LLM agents can dynamically draw on both sources to tackle complex, open-ended tasks~\cite{li2025evidence, miao2025recode}.

\begin{figure}[t]
    \centering
\includegraphics[width=1\linewidth,trim={0 0 129mm 20mm},clip]{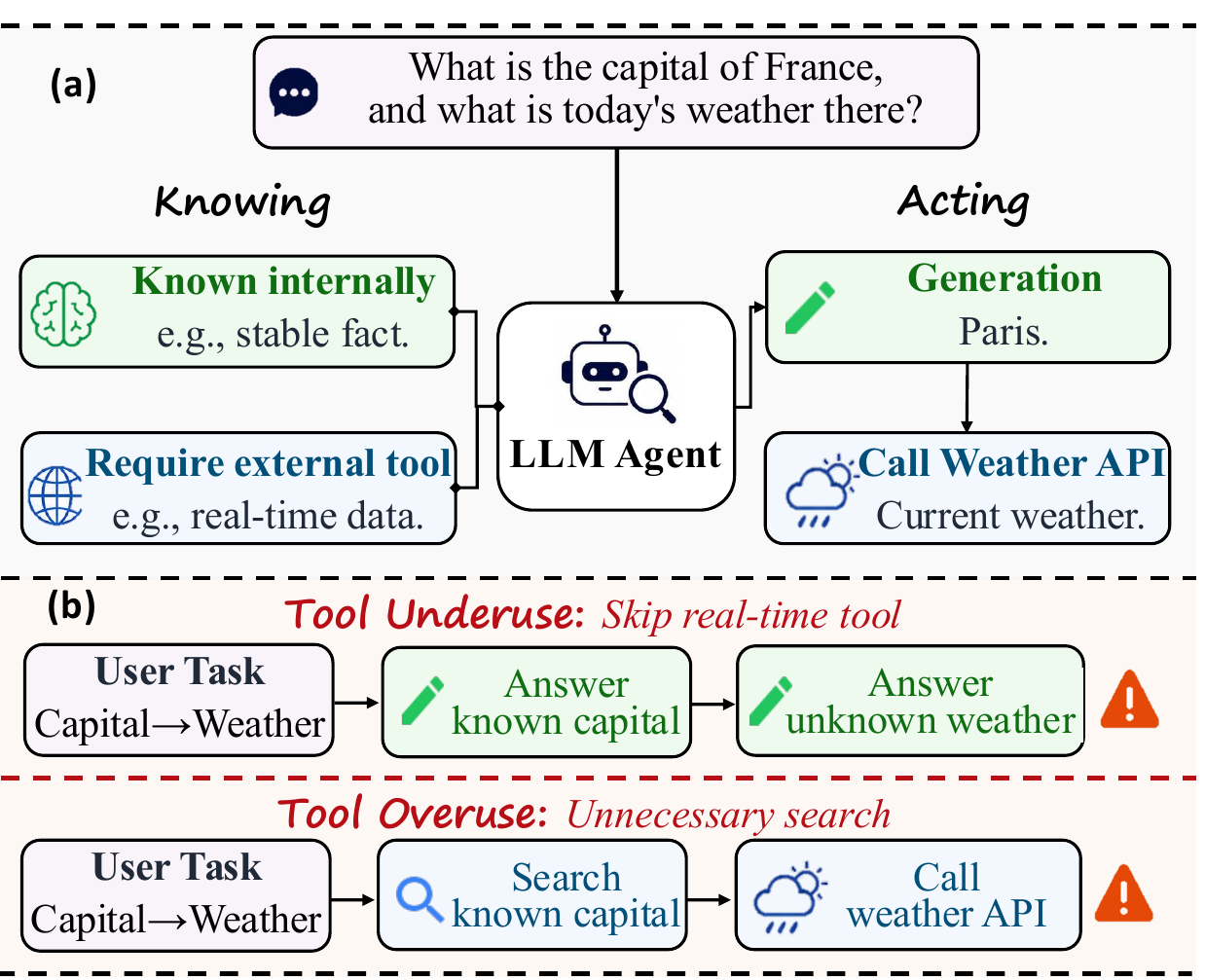}
    \caption{An illustration of self-awareness capability in LLM agents. (a) For a hybrid query (\textit{``capital of France'' + ``today's weather''}), an agent is expected to answer the capital from internal knowledge and then invokes the weather API. 
(b) Two failure modes: \textit{tool underuse} skips a necessary external call and hallucinates, while \textit{tool overuse} redundantly queries tools for known facts.}
    \label{fig:intro}
\end{figure}

Yet this duality poses a fundamental decision-making challenge: knowing when internal capabilities suffice, and when external tools are necessary.
Consider the hybrid task in Figure~\ref{fig:intro}(a): \textit{``What is the capital of France, and what is today's weather there?''} An ideal agent should answer the first part from parametric knowledge and invoke a weather API only for the second.
Failing to make this distinction leads to two characteristic failure modes:
\textbf{(1) tool underuse}, where the agent bypasses a necessary tool and falls back on hallucinated or outdated parametric knowledge.
\textbf{(2) tool overuse}, where the agent redundantly queries tools for facts already encoded in its parameters, inflating latency, cost, and noise.
Both modes reflect the same underlying deficiency: an inability to accurately assess one's own capability boundaries. We term this \textbf{self-awareness capability}: distinguish between what can be solved internally and what genuinely requires external tools, and to act accordingly~\cite{wang2025toward}.

However, existing benchmarks~\cite{qin2024toollm,liu-etal-2025-mcpeval,shen-etal-2024-taskbench} primarily emphasize task success in terms of task scale, complexity, and execution coverage.
This \textit{success-centric} paradigm treats tool invocation as an always-positive means to an end: as long as the task is completed, it does not matter whether the agent's decisions were appropriate. 
Consequently, such self-awareness capability is left systematically unmeasured. 
More critically, conflating outcome with process obscures a meaningful distinction: the agent does not know, or the agent knows but acts wrongly. 
These two cases call for different diagnoses, yet existing benchmarks conflate them. This raises a concrete question: 
\textbf{Do LLM agents actually possess self-awareness in tool-use decision making, and to what degree?}

From a cognitive perspective, \textit{true agency implies a unity where knowledge serves as the intentional foundation for action, and action acts as the strategic realization of knowledge}~\cite{prinz1997perception}. Guided by this, we introduce {\textit{\textbf{KAP}\scalebox{0.8}{\textbf{RO}}}} (\textit{K}nowing–\textit{A}cting Quadrant P\textit{RO}be), a framework that evaluates self-awareness by explicitly decoupling two dimensions: \textbf{(1) Knowing} is defined as the agent's explicit metacognitive judgment of its own epistemic boundaries, elicited by constraining direct execution and forcing the model to introspect on the necessity of external aid. \textbf{(2) Acting} represents the agent's spontaneous behavioral deployment observed within a standard, unconstrained tool-use environment.
 By juxtaposing these two dimensions, we can isolate \textbf{cognitive-behavioral alignment}: determining whether an agent’s failure stems from a genuine epistemic deficit (Unknown Unknowns) or a failure in executive control (Knowing but acting irrationally).

To operationalize this, we construct {\textit{\textbf{KA}\scalebox{0.8}{\textbf{ware}}}} dataset, which partitions the task space into three orthogonal subspaces defined by the interplay between parametric knowledge and information requirements. 
\textbf{External Function} targets strict external dependencies (\textit{e.g.}, real-time data, multimodal context).
\textbf{Internal Function} conversely focuses on tasks fully covered by internal capabilities (\textit{e.g}., translation, summarization).  \textbf{Hybrid Composition} involves dynamic complexity necessitating a divide-and-conquer strategy to discern which sub-steps require external versus internal processing. 
All tasks are synthesized by annotating tool seeds as external or internal attributes, then varying tool cardinality (single or multiple) and scenario complexity (single-hop, multi-hop, and parallel), enabling controlled and systematic probing.

Benchmarking a diverse set of agents spanning open-source and proprietary models, as well as instruction-following and reasoning-augmented architectures, yields three key findings. (1) Self-awareness exhibits a strong positive correlation with task success, underscoring its practical importance beyond a theoretical construct.
(2) Proprietary and reasoning-oriented models demonstrate more calibrated gating behavior, selectively abstaining from tool use when internal capabilities suffice, whereas open-source models tend toward pattern-driven invocation, triggering tools simply because they are available. 
(3) Across all models, self-awareness degrades substantially for tasks relying on internal parametric capabilities, revealing that internal capability calibration remains a shared and underappreciated weakness. Together, these results indicate that robust agent performance requires not only execution competence, but also reliable self-awareness capability.

\section{Related Work}
\paragraph{Agent Tool-Use Benchmarks.}
The shift from \emph{chat-only} large language models (LLMs) to \emph{tool-using} agents has motivated a growing set of benchmarks~\cite{li-etal-2023-api,qin2024toollm,patil2025bfcl,yao2025taubench}.
Early efforts annotate tool-use dialogues with API-call traces~\cite{li-etal-2023-api, huang-etal-2024-planning-creation}.
Complementary lines diagnose tool utilization at finer granularity: T-Eval decomposes sub-abilities~\cite{chen-etal-2024-eval},  and MCPEval extend evaluation to Model Context Protocol ecosystem~\cite{liu-etal-2025-mcpeval}.
Despite this diversity, existing benchmarks mainly reward \emph{successful tool invocation}, and provide limited supervision for the \emph{decision to refrain} from tool use.
Our work complements these benchmarks by explicitly formalizing the \emph{boundary} between internal capability and external tool need; Table~\ref{tab:differences_rebuttal} contrasts our benchmark with representative tool-use benchmarks along four axes, where only ours satisfies all.

\paragraph{LLM Agent Knowledge Boundary.}
Deciding whether to \emph{act} via tools or \emph{answer} from parametric knowledge is closely tied to an agent's knowledge boundary. 
Self-RAG demonstrate that retrieval invocation should be conditional on need~\cite{asai2023selfrag}, and MetaTool evaluates whether and which tools to use~\cite{huang2023metatool}.
Latest work further frames agents as tool-use decision-makers and argues that the decision boundary should be aligned with the knowledge boundary, focusing on mitigating \emph{tool overuse}~\cite{wang2025toward,wu2026callframeworkassessoptimize,zeng2026tooloveruseillusiondoesllm}.
However, these efforts target overuse in isolation and rarely release a self-awareness evaluation protocol.
We close this gap by partitioning the task space, turning a single benchmark into a unified probe of both miscalibration directions, and extend to both tool overuse and underuse.
More discussion of prior work is provided in Appendix~\ref{appendix:relatedworks}.

 \begin{table}[h]
\centering
\scriptsize
\setlength{\tabcolsep}{5pt}
\caption{Comparison to representative benchmarks.}\label{tab:differences_rebuttal}
\vspace{-0.2cm}
\begin{tabular}{@{}lcccc@{}}
\toprule
Benchmark & \# MCPs   & Logic Ann. & Outcome Eval. & Boundary Eval.\\ \midrule
ToolBench & 0& \xmark & \xmark & \xmark \\
$\tau$-Bench & 0 & \xmark & \cmark & \xmark \\
MCP-Bench & 28 & \cmark & \cmark & \xmark  \\
\textbf{Ours} & 45 & \cmark & \cmark & \cmark \\
\bottomrule
\end{tabular}
\vspace{-0.3cm}
\end{table}
\section{{\textit{\textbf{KA}\scalebox{0.8}{\textbf{ware}}}}: A Benchmark for  Capability Self-awareness }
This section details the construction and evaluation pipeline, covering tool seed construction, task synthesis, and the \textit{{KAP}\scalebox{0.8}{{RO}}} evaluation protocol.

\subsection{Formulation}

To evaluate the self-awareness of LLM agents, we partition the capabilities required for a task into two disjoint sets: (1) \emph{Parametric Capability} ($\mathcal{C}_{\text{param}}$), inherent to the static model parameters $\theta$. 
(2) \emph{Tool-dependent Capability} ($\mathcal{C}_{\text{tool}}$), which strictly requires interaction with external toolset $\mathcal{T}$.
Let $\mathcal{Q}$ denote the universe of user queries. Each query $q \in \mathcal{Q}$ is decomposed into a set of subtasks $S(q) = \{s_1, \ldots, s_m\}$, where $\Phi(s)$ denotes the capabilities necessary to resolve subtask $s$. We define a capability boundary indicator $f_B : \mathcal{S} \rightarrow \{0,1\}$ as:
\begin{equation}
f_B(s) = \mathbf{I}\!\big(\Phi(s) \cap \mathcal{C}_{\text{tool}} \neq \emptyset\big),
\end{equation}
where $f_B(s)=1$ implies subtask $s$ strictly requires external tools, and $f_B(s)=0$ implies it is solvable via internal parametric inference. The tool requirement pattern of a task is then:
\begin{equation}
\mathbf{b}(q) = \big(f_B(s_1), \ldots, f_B(s_m)\big).
\end{equation}
This naturally characterizes three subspaces:
\textit{\textbf{External Function}} ($\mathbf{b}(q)=\mathbf{1}$),
\textit{\textbf{Internal Function}} ($\mathbf{b}(q)=\mathbf{0}$),
and \textit{\textbf{Hybrid Composition}} (there exist $i,j$ such that $f_B(s_i)=1$ and $f_B(s_j)=0$).

\subsection{Tool Seed Construction}
We begin by building a high-quality tool seed pool. We employ a cascading pipeline that denoises and structures the vast MCP tool space. 
Per-step statistics and prompts are provided in Appendix~\ref{appendix:tool evaluation}.

\begin{figure*}[t]
    \centering
\includegraphics[width=1\linewidth]{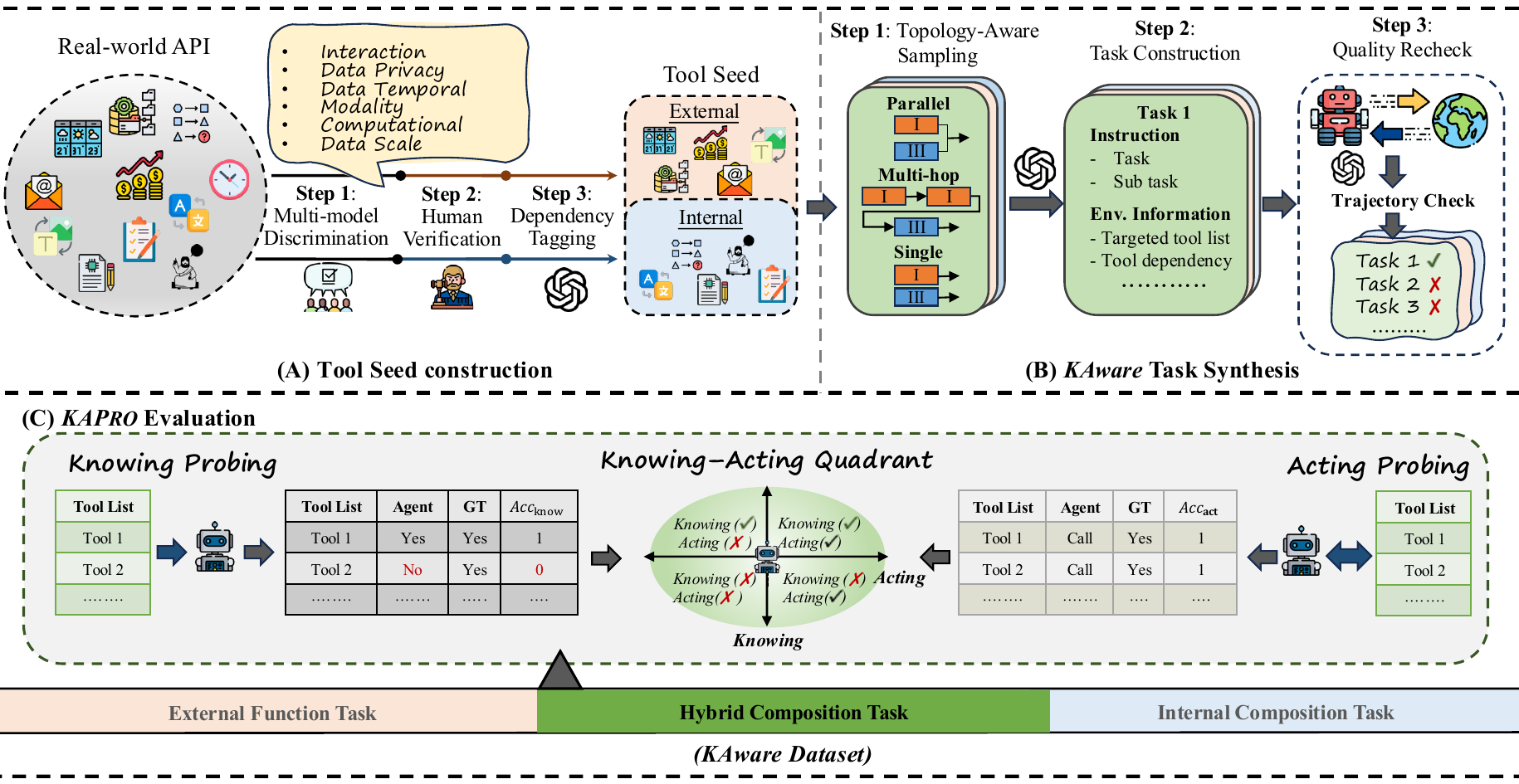}
    \caption{An illustration of {\textit{\textbf{KAP}\scalebox{0.8}{\textbf{RO}}}} pipeline. \textbf{(A)} \textbf{Tool Seed construction}: Real-world APIs are filtered and annotated into \textit{\textbf{External Function}} and \textit{\textbf{Internal Function}} tool seeds.
\textbf{(B)} {\textit{\textbf{KA}\scalebox{0.8}{\textbf{ware}}}} \textbf{Task Synthesis}:  we utilize topology-aware sampling (parallel, multi-hop, single) to construct diverse tasks, followed by a rigorous quality recheck of agent trajectories.
\textbf{(C) }{\textit{\textbf{KAP}\scalebox{0.8}{\textbf{RO}}}} \textbf{Evaluation}: we assess knowing-acting alignment by decoupling \textit{Knowing} (metacognitive judgment of tool necessity) from \textit{Acting} (spontaneous behavioral execution).}
    \label{fig:model}
\end{figure*}

\paragraph{Tool Discrimination.}
We  aggregate tool operations from diverse benchmarks (e.g., ToolBench~\cite{qin2024toollm}, $\tau$-bench~\cite{yao2025taubench}, Toucan~\cite{xu2025toucan}) into a unified schema $\mathcal{S}_{tool} = \langle \text{name, params, description} \rangle$.
To automate the assessment of $t \in \mathcal{T}$, we establish a six-dimensional rubric: \textit{Interaction}     
  (state-mutating operations), \textit{Temporality} (real-time data), \textit{Data Privacy} (authenticated information), \textit{Data Scale}     
  (exceeding context limits), \textit{Computational} (complex computations), and \textit{Modality} (non-native modality). 
 Each dimension captures a distinct capacity that LLMs inherently lack. 
A tool is labeled \textit{External} if it triggers any dimension, and \textit{Internal} only if it triggers none, yielding 1,286 internal and 26,616 external operations (Appendix~\ref{appendix:tool statistics}).

 \paragraph{Human Verification.}
  To audit whether the automated  labels are reliable, three annotators independently re-label a stratified subset of 400 tools using the same six-dimensional rubric. The  
  final human label is determined by majority vote. The human--automated agreement reaches 90.41\%, confirming the reliability (details in Appendix~\ref{appendix:tool statistics}).

\paragraph{Dependency Tagging.}
Isolated tools are insufficient for complex tasks, tools deployed in similar workflows (e.g., \textit{flight booking} and \textit{hotel reservation}) are far more likely to be co-invoked than tools from disjoint domains. We therefore surface candidate combinations via a usage-scenario prior: for each tool $t \in \mathcal{T}$, we prompt an LLM to synthesize a representative usage scenario $\rho_t$, and score a pair $(t_i, t_j)$ by the cosine similarity of their scenario embeddings:
\begin{equation}
    \text{Affinity}(t_i, t_j) = \cos\!\big(\mathbf{E}(\rho_{t_i}),\, \mathbf{E}(\rho_{t_j})\big),
\end{equation}
where $\mathbf{E}(\cdot)$ denotes the embedding function (we use \textit{bge-large-en-v1.5}). $\text{Affinity}$ serves as a soft prior rather than a hard dependency, and the executable dependency is then established and validated in the subsequent task synthesis stage.

\subsection{Task Synthesis}
\label{subsec:synthesis_pipeline}
We adopt a three-stage pipeline of \emph{sampling}, \emph{generation}, and \emph{execution} to generate tasks that approximate the complexity of real user queries.

\paragraph{Topology-Aware Sampling.}
We formalize task generation as sampling from a conditional probability distribution $P(\text{Task} \mid \mathcal{T}_{sub}, \mathcal{C})$, where $\mathcal{T}_{sub} \subseteq \mathcal{T}$ is the sampled tool subset of size $k = |\mathcal{T}_{sub}| \in \{1,2,3\}$, and $\mathcal{C}$ denotes the target complexity (e.g., single-hop, parallel, or multi-hop).
For multi-tool scenarios ($k \geq 2$), we sample $\mathcal{T}_{sub}$ to maximize the joint affinity score, encouraging the sampling toward semantically coherent combinations.

\paragraph{Task Construction and Quality Check.} 
Given $\mathcal{T}_{\text{sub}}$, a generator $\mathcal{M}_{\text{gen}}$ synthesizes a pair $\langle q, \mathcal{P}^* \rangle$. Here, $q$ is the natural language user query, and $\mathcal{P}^*$ represents the ground truth plan, comprising the dependency graph $G_{dep}$ and the hierarchical sub-task decomposition.
Each candidate first passes a rule-based check (schema validity, parameter bindability, and tool-name resolvability), and is then judged by an LLM evaluator on six binary dimensions: \emph{knowledge boundary} ($q$ exceeds parametric knowledge, e.g., real-time or private data), \emph{selection difficulty} (the tool is not surface-named in $q$), \emph{tool uniqueness} (no equivalent substitute in $\mathcal{T}_{\text{sub}}$), \emph{parameter completeness} (required arguments are grounded in $q$), \emph{scenario realism} ($q$ reflects a plausible workflow), and \emph{answer verifiability} (correctness is objectively verifiable). Any sample failing a single criterion is discarded. The full evaluation prompt and per-dimension scoring rubric are deferred to Appendix~\ref{appendix:task gen and quality}.

\paragraph{Trajectory Acquisition and Quality Recheck}
Filtered tasks are instantiated in an executable agent environment. A strong agent $\mathcal{A}$ (gpt-5) autonomously plans and invokes tools conditioned on $\langle q, G_{\text{dep}}, \mathcal{T}_{\text{sub}} \rangle$, producing a trajectory $\tau = [(a_1, o_1), \dots, (a_T, o_T)]$, where $a_t$ denotes an action (inference or tool invocation) and $o_t$ the environmental observation. 
Each trajectory $\tau$ is then re-assessed by an LLM evaluator on three binary dimensions: \emph{causal consistency} (each action $a_t$ is grounded in prior observations $o_{<t}$ and respects the dependency graph $G_{\text{dep}}$, with no unmotivated tool calls), \emph{boundary consistency} (the realized tool-use pattern of $\tau$ matches the prescribed capability pattern), and \emph{execution accuracy} (all tool calls return successful responses and the final answer matches the ground-truth derived from $\mathcal{P}^*$). Only trajectories passing all three criteria are retained.

\subsection{Data Statistics}
  The benchmark comprises 1,076 tasks spanning three capability boundaries and three logic types  (Table~\ref{tab:dataset_stats}).
  As an independent quality check, three annotators re-inspect 141 stratified instances,
  verifying task descriptions, logic-type annotations, and tool dependency structures. The human agreement
  reaches 96.45\% overall (\textit{External Function} 100\%, \textit{Hybrid Composition} 90.91\%, \textit{Internal Function} 97.92\%). Details can be found in
  Appendix~\ref{appendix: dataset Human Evaluation}.

  \begin{table}[h]
    \centering
    \resizebox{\linewidth}{!}{
        \begin{tabular}{lcccc}
            \toprule
            \textbf{Task Type} & \textbf{Single-hop} & \textbf{Multi-hop} & \textbf{Parallel} & \textbf{Total} \\
            \midrule
            \textit{\textbf{External Function}}  & 106 & 131 & 73  & 310 \\
            \textit{\textbf{Hybrid Composition}}  &  -  & 311 & 57  & 368 \\
            \textit{\textbf{Internal Function}}   & 138 & 130 & 130 & 398 \\
            \bottomrule
        \end{tabular}
    }
    \caption{Dataset statistics by task and logical type}
    \label{tab:dataset_stats}
\end{table}


\subsection{{\textit{\textbf{KAP}\scalebox{0.8}{\textbf{RO}}}}  Evaluation}
\paragraph{Probing Settings}
We design two complementary probing settings that decouple metacognitive judgment from behavioral execution.

\textit{Knowing Probing: }This setting evaluates the agent's ability to recognize its own cognitive boundaries. By restricting direct execution, we force the model to reflect on the necessity of external tools. For each task $d_i$, the agent explicitly predicts the set of needed tools, denoted as $E_{\mathrm{know}}^{(i)}$. 

\textit{Acting Probing:} This setup tests the agent's autonomous tool deployment in a standard environment, where it self-decides whether to invoke tools to solve unconstrained problems. We record $E_{\mathrm{act}}^{(i)}$, which denotes the set of tools actually invoked during the execution of task $d_i$.  
In addition, we report \textit{Pass Rate} for end-to-end correctness, using gpt-4.1 as a judge that assigns $1$ if the trajectory and final answer are both correct and $0$ otherwise (details in Appendix~\ref{appendix:experiment}).

\begin{table*}[t]
    \centering
    \resizebox{\textwidth}{!}{%
        \begin{tabular}{l|cccc|cccc|cccc|c}
            \toprule
            \multirow{2}{*}{\textbf{Model}} & \multicolumn{4}{c|}{\textit{\textbf{External Function}}} & \multicolumn{4}{c|}{\textit{\textbf{Hybrid Composition}}} & \multicolumn{4}{c|}{\textit{\textbf{Internal Function}}} & {\textbf{Avg}} \\
            \cmidrule(lr){2-8} \cmidrule(lr){6-9} \cmidrule(lr){10-14} &\textbf{Pass}
             & \textbf{$\text{Acc}_{\text{know}}$} & \textbf{$\text{Acc}_{\text{act}}$} & \textbf{KAS} & \textbf{Pass} & \textbf{$\text{Acc}_{\text{know}}$} & \textbf{$\text{Acc}_{\text{act}}$} & \textbf{KAS} & \textbf{Pass} & \textbf{$\text{Acc}_{\text{know}}$} & \textbf{$\text{Acc}_{\text{act}}$} & \textbf{KAS} & \textbf{KAS}\\
            \midrule

            \multicolumn{14}{c}{\textit{\textbf{Closed-source Models}}} \\
            \midrule
            \rowcolor[gray]{.95} \multicolumn{14}{c}{\textit{Instruct Models}} \\
            gpt-4o & 85.16 & \textbf{99.41} & 83.87 & 90.98 & 65.49 & 53.94 & 37.18 & 44.02 & 59.30 & 24.37 & 46.48 & 31.98 & 53.10 \\
            gpt-4.1 & 87.10 & 99.03 & 84.62 & 91.26 & 64.95 & 55.03 & 37.05 & 44.28 & 59.05 & 27.39 & 52.43 & 35.98 & 54.75 \\
            gemini-3-flash & 85.81 & 99.35 & 81.60 & 89.61 & 64.95 & 60.33 & 36.28 & 45.31 & 59.05 & 27.89 & 41.12 & 33.24 & 53.61 \\
            claude-sonnet-4.5-instruct & 93.55 & 98.87 & 92.58 & 95.62 & 74.73 & 51.49 & 44.29 & 47.62 & 87.94 & 29.15 & 22.99 & 25.70 & 53.34 \\
            qwen3-max & 97.10 & 98.82 & 95.94 & 97.36 & \textbf{97.01} & 51.18 & 55.43 & 53.22 & \textbf{95.98} & 13.32 & 18.59 & 15.52 & 51.99 \\
            \cdashline{1-14}[0.5pt/2pt]
            \rowcolor[gray]{.95} \multicolumn{14}{c}{\textit{Reasoning Models}} \\
            gpt-5 & 84.19 & 96.56 & 81.45 & 88.36 & 63.32 & \textbf{85.96} & 36.59 & 51.33 & 60.80 & \textbf{94.22} & \textbf{58.54} & \textbf{72.22} & \textbf{69.73} \\
            o4-mini & 84.52 & 96.88 & 82.69 & 89.22 & 63.32 & 69.34 & 39.49 & 50.32 & 60.05 & 60.05 & 54.98 & 57.41 & 64.15 \\
            gpt-5.5 & 95.81 & 97.96 & 93.92 & 95.90 & 84.78 & 85.51 & 50.05 & 63.14 & 86.68 & 68.09 & 32.66 & 44.15 & 65.55 \\
            gemini-3-pro & 88.06 & 98.01 & 86.24 & 91.75 & 88.86 & 74.14 & \textbf{57.29} & \textbf{64.64} & 58.79 & 36.68 & 41.25 & 38.83 & 62.90 \\
        claude-sonnet-4.5-think & 97.71 & 98.39 & 96.24 & 97.30 & 98.97 & 61.77 & 55.76 & 58.61 & 96.23 & 44.72 & 14.45 & 21.84 & 55.68 \\
                    \midrule
            \multicolumn{14}{c}{\textit{\textbf{Open-source Models}}} \\
            \midrule
            \rowcolor[gray]{.95} \multicolumn{14}{c}{\textit{Instruct Models}} \\
            qwen3-235b-instruct & 63.87 & 98.71 & 57.26 & 72.48 & 86.96 & 53.44 & 43.95 & 48.23 & 90.45 & 8.04 & 45.48 & 13.66 & 42.43 \\
            qwen3.5-397b-instruct & 96.13 & 98.87 & 94.46 & 96.62 & 88.32 & 57.56 & 45.97 & 51.12 & 84.67 & 39.45 & 21.86 & 28.13 & 55.72 \\
            deepseek-v3.2 & 83.55 & 98.71 & 84.14 & 90.84 & 77.45 & 52.81 & 46.15 & 49.26 & 86.18 & 14.07 & 20.06 & 16.54 & 49.14 \\
            deepseek-v4-flash & 92.26 & 98.28 & 93.82 & 96.00 & 85.60 & 57.52 & 46.78 & 51.60 & 83.42 & 33.17 & 16.33 & 21.89 & 53.40 \\
            \cdashline{1-14}[0.5pt/2pt]
            \rowcolor[gray]{.95} \multicolumn{14}{c}{\textit{Reasoning Models}} \\
            qwen3-235b-think & \textbf{98.04} & 97.65 & \textbf{97.19} & 97.42 & 83.89 & 54.49 & 45.42 & 49.54 & 93.22 & 24.62 & 18.93 & 21.40 & 50.54 \\
            qwen3.5-397b-think & 93.87 & 98.87 & 95.86 & 97.34 & 93.75 & 57.70 & 49.73 & 53.42 & 84.92 & 39.95 & 22.36 & 28.67 & 56.92 \\
            deepseek-r1 & 53.87 & 98.60 & 55.91 & 71.36 & 62.77 & 67.21 & 36.01 & 46.90 & 58.54 & 40.95 & 31.62 & 35.68 & 49.80 \\
            deepseek-v4-pro & 97.42 & 98.55 & 97.10 & \textbf{97.82} & \textbf{97.01} & 57.93 & 53.35 & 55.54 & 83.42 & 32.91 & 17.34 & 22.71 & 55.58 \\

            \bottomrule
        \end{tabular}
    }

    \caption{Comprehensive evaluation of model performance across three settings: \textit{External Function}, \textit{Hybrid Composition}, and \textit{Internal Function}. We report Pass Rate, $\text{Acc}_{\text{know}}$, $\text{Acc}_{\text{act}}$, KAS. KAS exhibits a negative correlation the task's reliance on internal knowledge, indicating that ambiguity regarding the necessity of tools (i.e., when tools are available but not required) significantly impairs the self-awareness capability.}
    \label{tab:comprehensive_model_performance}
\end{table*}

\paragraph{Evaluation Metrics.}
Let the evaluation set be $D = \{d_1, \dots, d_N\}$, and let $T^{(i)}$ denote the ground-truth tools required by task $d_i$. For any tool set $X$, we measure tool-set alignment with the reference set $T$ via Jaccard similarity:
\begin{equation}
J(X,T)=\frac{|X \cap T|}{|X \cup T|},
\end{equation}
where $J(X,T)=1$ when $X=T=\emptyset$. 
The \emph{knowing accuracy} and \emph{acting accuracy} are then calculated as:
\begin{equation}
\mathrm{Acc}_{\mathrm{know}} = \frac{1}{|D|} \sum_{i=1}^{|D|} J(E_{\mathrm{know}}^{(i)}, T^{(i)}),
\end{equation}
\begin{equation}
\mathrm{Acc}_{\mathrm{act}} = \frac{1}{|D|} \sum_{i=1}^{|D|} J(E_{\mathrm{act}}^{(i)}, T^{(i)}).
\end{equation}
To distinguish directional tool-use errors, we further report tool overuse and underuse for both knowing and acting; their set-based definitions and Jaccard-error decomposition are provided in Appendix~\ref{appendix:metrics}.
To jointly assess capability awareness at both levels, we aggregate the two via their harmonic mean, yielding the 
\emph{Know-Act Joint Jaccard Score (KAS)}:
\begin{equation}
\mathrm{KAS} = 2 \left( \mathrm{Acc}_{\mathrm{act}}^{-1} + \mathrm{Acc}_{\mathrm{know}}^{-1} \right)^{-1},
\end{equation}
where \emph{KAS} $=0$ whenever either component accuracy is zero. \emph{KAS} is high only when the agent exhibits a balanced capability in both  \emph{knowing} tool necessity and \emph{acting} upon it, thereby penalizing the overall gap between cognition and behavior. Appendix~\ref{appendix:metrics} further justifies the design principles.

\section{Experiments}
\subsection{Setups}
\paragraph{Models}
We evaluate 18 representative LLMs, including
 gpt-4o~\cite{hurst2024gpt}, gpt-4.1~\cite{gpt4_1}, o4-mini~\cite{o4_mini_20250416}, gpt-5~\cite{singh2025openai}, claude-sonnet-4.5~\cite{claude_sonnet_45}, gemini-3-flash~\cite{gemini3_flash_preview}, gemini-3-pro~\cite{gemini3_pro_preview} and qwen3-max~\cite{qwen3max}. We further include open-source models: qwen3-235b-a22b~\cite{qwen3technicalreport}, qwen3.5-397b-a17b~\cite{qwen3.5}, deepseek-v3.2~\cite{deepseekai2025deepseekv32}, deepseek-r1~\cite{deepseekai2025deepseekr1incentivizingreasoningcapability} and deepseek-v4~\cite{deepseekai2026deepseekv4}.
The details of models are reported in Appendix~\ref{appendix:api details}.

\paragraph{Implementation Details.}
We evaluate all models zero-shot with greedy decoding. Closed-source models are queried via
their official APIs. We report Pass Rate by using gpt-4.1 to judge the correctness of the
agent's final answer.
Details of evaluation settings are provided in the Appendix~\ref{appendix:Implementation Details}.

\subsection{Main Results}

\begin{table*}[htbp]
    \centering
    \resizebox{\textwidth}{!}{%
    \begin{tabular}{l|ccccc|ccccc|ccccc|c}
        \toprule
        \multirow{2}{*}{\textbf{Model}} & \multicolumn{5}{c|}{\textit{\textbf{External Function}}} & \multicolumn{5}{c|}{\textit{\textbf{Hybrid Composition}}} & \multicolumn{5}{c|}{\textit{\textbf{Internal Function}}} & \textbf{Avg} \\
        \cmidrule(lr){2-7} \cmidrule(lr){7-11} \cmidrule(lr){12-17}
         & \textbf{Kc/Ac} & \textbf{Kc/Aw} & \textbf{Kw/Ac} & \textbf{Kw/Aw} & \textit{\textbf{DirGap}} & \textbf{Kc/Ac} & \textbf{Kc/Aw} & \textbf{Kw/Ac} & \textbf{Kw/Aw} & \textit{\textbf{DirGap}} & \textbf{Kc/Ac} & \textbf{Kc/Aw} & \textbf{Kw/Ac} & \textbf{Kw/Aw} & \textit{\textbf{DirGap}} &  \textit{\textbf{DirGap}}\\
        \midrule
        
        \multicolumn{17}{c}{\textit{\textbf{Closed-source Models}}} \\
        \midrule
        \rowcolor[gray]{.95} \multicolumn{17}{c}{\textit{Instruct Models}} \\
        gpt-4o & 79.03 & 19.35 & 0.97 & 0.65 & $+$18.38 & 0.82 & 4.89 & 4.89 & 89.40 & $+$0.00 & 14.57 & 9.80 & 31.91 & 43.72 & $-$22.11 & $-$2.88 \\
        gpt-4.1 & 80.32 & 17.74 & 0.65 & 1.29 & $+$17.09 & 1.63 & 6.79 & 4.89 & 86.68 & $+$1.90 & 19.60 & 7.79 & 32.41 & 40.20 & $-$24.62 & $-$3.53 \\
        gemini-3-flash & 75.16 & 23.23 & 0.97 & 0.65 & $+$22.26 & 1.09 & 17.39 & 4.62 & 76.90 & $+$12.77 & 9.80 & 18.09 & 30.90 & 41.21 & $-$12.81 & $+$6.04 \\
        claude-sonnet-4.5-instruct & 87.74 & 10.32 & 0.32 & 1.61 & $+$10.00 & 0.27 & 0.82 & 6.79 & 92.12 & $-$5.97 & 6.03 & 23.12 & 14.57 & 56.28 & $+$8.55 & $+$4.00 \\
        qwen3-max & 90.65 & 6.45 & 2.26 & 0.65 & $+$4.19 & 0.27 & 0.27 & 11.96 & 87.50 & $-$11.69 & 6.03 & 7.29 & 12.06 & 74.62 & $-$4.77 & $-$4.56 \\
        \cdashline{1-17}[0.5pt/2pt]
        \rowcolor[gray]{.95} \multicolumn{17}{c}{\textit{Reasoning Models}} \\
        gpt-5 & 71.94 & 20.65 & 4.84 & 2.58 & $+$15.81 & 7.88 & 73.64 & 1.09 & 17.39 & $+$72.55 & 55.28 & 38.94 & 3.27 & 2.51 & $+$35.67 & $+$42.56 \\
        o4-mini & 75.16 & 19.03 & 2.58 & 3.23 & $+$16.45 & 6.52 & 30.98 & 8.15 & 54.35 & $+$22.83 & 35.93 & 24.12 & 17.59 & 22.36 & $+$6.53 & $+$14.96 \\
        gpt-5.5 & 86.45 & 9.35 & 3.23 & 0.97 & $+$6.12 & 13.59 & 57.34 & 1.36 & 27.72 & $+$55.98 & 22.61 & 45.48 & 10.05 & 21.86 & $+$35.43 & $+$34.01 \\
        gemini-3-pro & 79.03 & 16.45 & 2.90 & 1.61 & $+$13.55 & 6.52 & 41.85 & 13.32 & 38.32 & $+$28.53 & 14.32 & 22.36 & 26.63 & 36.68 & $-$4.27 & $+$12.08 \\
        claude-sonnet-4.5-think & 90.97 & 6.13 & 2.26 & 0.65 & $+$3.87 & 9.28 & 13.40 & 27.32 & 50.00 & $-$13.92 & 2.76 & 41.96 & 8.29 & 46.98 & $+$33.67 & $+$7.87 \\
        \midrule
        
        \multicolumn{17}{c}{\textit{\textbf{Open-source Models}}} \\
        \midrule
        \rowcolor[gray]{.95} \multicolumn{17}{c}{\textit{Instruct Models}} \\
        qwen3-235b-instruct & 53.23 & 43.87 & 1.29 & 1.61 & $+$42.58 & 0.27 & 3.80 & 8.15 & 87.77 & $-$4.35 & 6.28 & 1.76 & 38.94 & 53.02 & $-$37.18 & $-$2.97 \\
        qwen3.5-397b-instruct & 88.71 & 8.39 & 1.61 & 1.29 & $+$6.78 & 1.63 & 11.96 & 2.99 & 83.42 & $+$8.97 & 12.31 & 27.14 & 9.55 & 51.01 & $+$17.59 & $+$11.53 \\
        deepseek-v3.2 & 78.39 & 18.71 & 1.94 & 0.97 & $+$16.77 & 1.90 & 2.45 & 8.70 & 86.96 & $-$6.25 & 1.76 & 12.31 & 17.84 & 68.09 & $-$5.53 & $+$0.65 \\
        deepseek-v4-flash & 86.13 & 10.32 & 2.58 & 0.97 & $+$7.74 & 1.63 & 11.68 & 4.08 & 82.61 & $+$7.60 & 5.28 & 27.89 & 11.06 & 55.78 & $+$16.83 & $+$11.05 \\
        \cdashline{1-17}[0.5pt/2pt]
        \rowcolor[gray]{.95} \multicolumn{17}{c}{\textit{Reasoning Models}} \\
        qwen3-235b-think & 92.16 & 4.31 & 2.35 & 1.18 & $+$1.96 & 2.22 & 4.17 & 13.33 & 80.28 & $-$9.16 & 4.77 & 19.85 & 13.57 & 61.81 & $+$6.28 & $-$0.25 \\
        qwen3.5-397b-think & 90.32 & 6.77 & 2.26 & 0.65 & $+$4.51 & 1.63 & 12.23 & 2.45 & 83.70 & $+$9.78 & 12.56 & 27.39 & 9.80 & 50.25 & $+$17.59 & $+$11.15 \\
        deepseek-r1 & 50.97 & 45.48 & 0.97 & 2.58 & $+$44.51 & 1.63 & 33.15 & 7.88 & 57.34 & $+$25.27 & 7.29 & 33.67 & 23.37 & 35.68 & $+$10.30 & $+$25.28 \\
        deepseek-v4-pro & 90.97 & 5.48 & 2.90 & 0.65 & $+$2.58 & 5.16 & 8.42 & 4.62 & 81.79 & $+$3.80 & 7.54 & 25.38 & 9.80 & 57.29 & $+$15.58 & $+$7.81 \\
        \bottomrule
    \end{tabular}%
    }
    \caption{Know-Act quadrant proportions across \textit{External Function}, \textit{Hybrid Composition}, and \textit{Internal Function} settings. Kc/Kw denote knowing correct/wrong; Ac/Aw denote acting correct/wrong. \textit{DirGap} $=p_{Kc/Aw}-p_{Kw/Ac}$ is the directional  gap between knowing and acting.}
    \label{table:Q1Q2Q3Q4}
\end{table*}

\begin{figure*}[t]
\centering
    \begin{subfigure}[b]{0.32\linewidth}
    \centering
      \includegraphics[width=\linewidth]{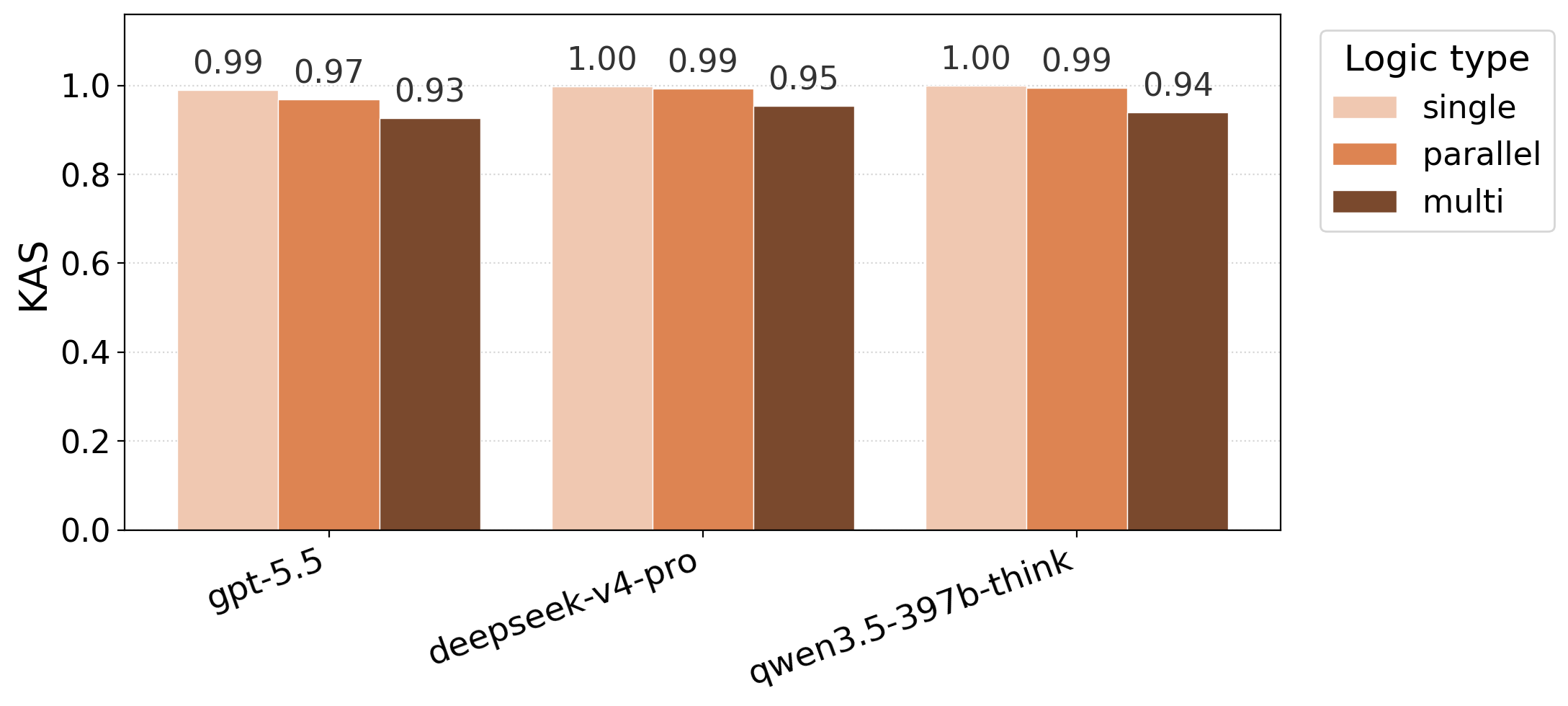}
      \subcaption{\textit{\textbf{External Function}}}
    \end{subfigure}
\hfill 
    \begin{subfigure}[b]{0.32\linewidth}
    \centering
    \includegraphics[width=\linewidth]{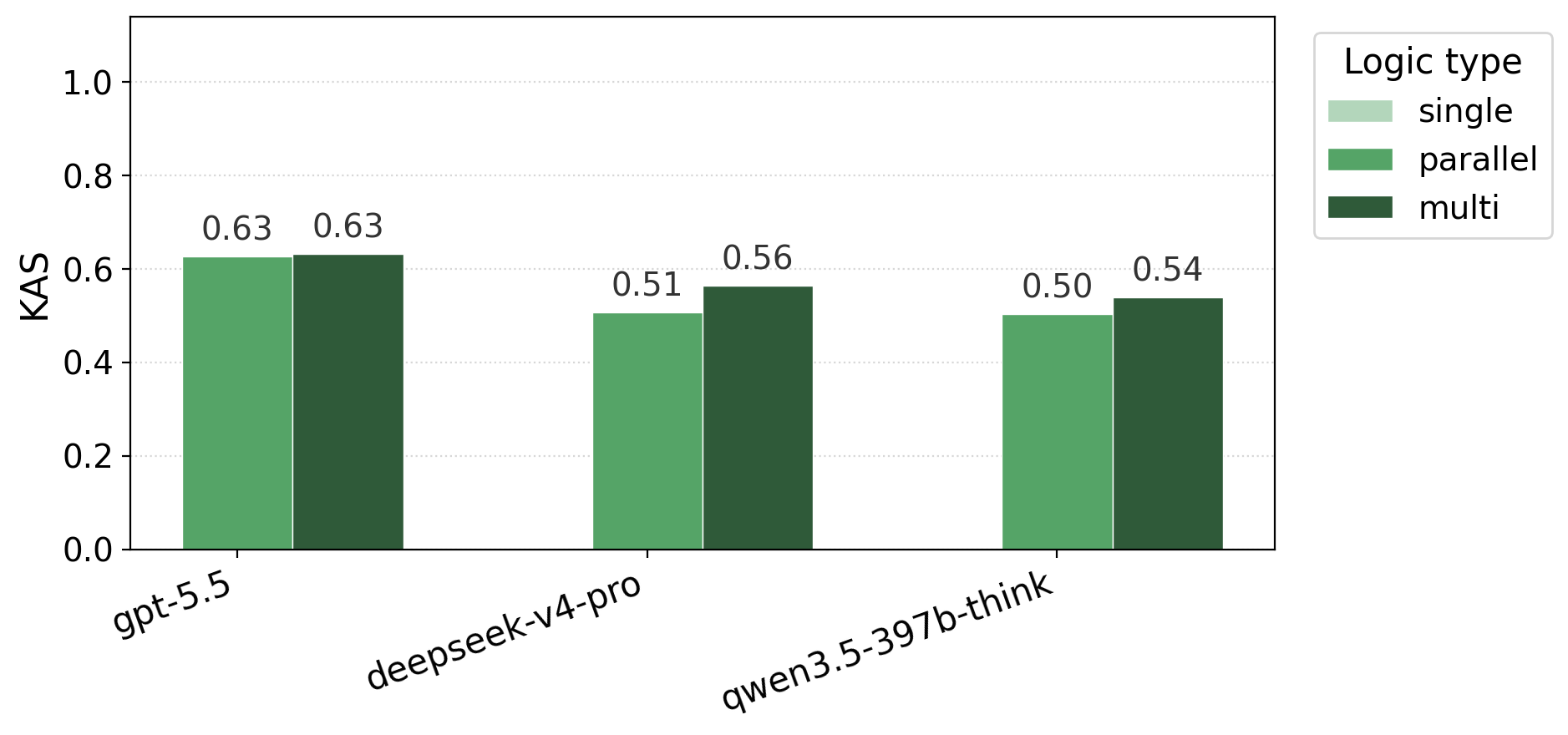}
      \subcaption{\textit{\textbf{Hybrid Composition}}}
    \end{subfigure}
\hfill
    \begin{subfigure}[b]{0.32\linewidth}
    \centering
      \includegraphics[width=\linewidth]{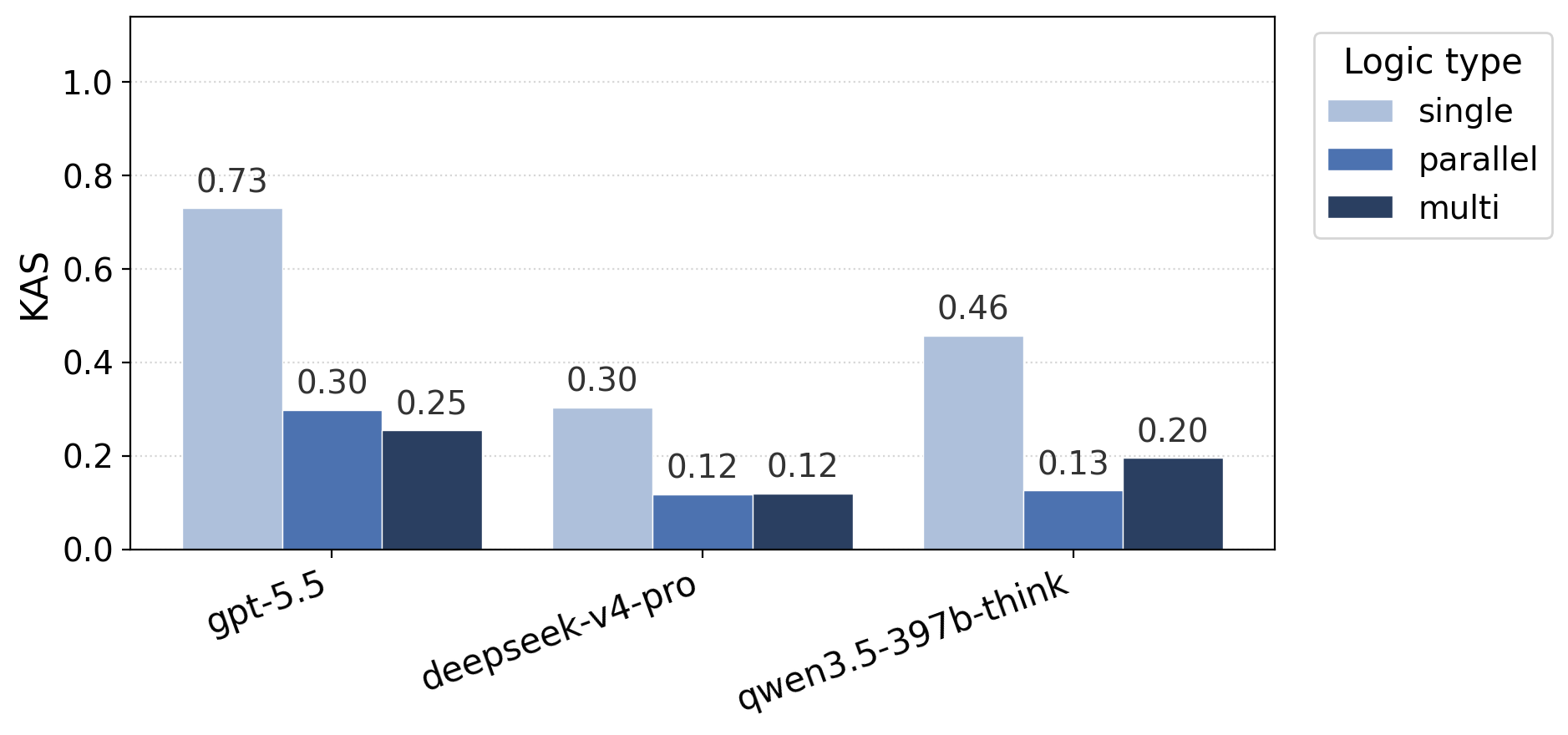}
      \subcaption{\textit{\textbf{Internal Function}} }
    \end{subfigure}

    \caption{KAS performance across logical reasoning types (Single-hop, Parallel, Multi-hop) and models. A systematic decline in KAS as combinatorial complexity increases, with sequential multi-hop reasoning causing more severe degradation than parallel reasoning.}
\label{fig:logic_types}
\end{figure*}

\paragraph{Pass Rate is insufficient for measuring self-awareness  capability, while KAS reveals deeper tool-use patterns.}
Table~\ref{tab:comprehensive_model_performance} shows that end-to-end task pass rate tightly tracks self-awareness only in the purely tool-dependent setting.
This alignment dissolves once internal knowledge is involved, and a high Pass Rate can mask poor self-awareness: in \textit{Internal Function}, a model can answer correctly from internal capability while still issuing unnecessary tool calls, and Pass Rate silently absorbs such tool overuse. Qwen3-max's 95.98 Pass Rate against only 15.52 KAS, revealing that high task success can coexist with severely miscalibrated tool decisions.
By grounding both the model's explicit judgment and its actual tool-use behavior, KAS captures the miscalibration that Pass Rate alone overlooks.

\paragraph{KAS reveals systematic advantages of reasoning and closed-source models.}
Reasoning models surpass their instruction-following variants on KAS (e.g., $+$9.3 in gemini-3).
This is consistent with the thinking stage acting as an internal verifier: before committing to action, reasoning models reassess whether the task truly exceeds their parametric capability.
Further, while open-source models are competitive on task pass rate (e.g., qwen3-235b-think tops \textit{{External Function}} Pass at 98.04), yet the highest KAS scores all belong to closed-source reasoning models, with the gap widening sharply in \textit{Internal Function}.
In both hierarchies, the gap localizes to the same underlying skill: knowing when to invoke tools and when not.

\paragraph{Model families exhibit distinct boundary-aware behaviors.}
The results also reveal notable differences across model
families. The \textit{GPT} family is the most balanced calibrator,  showing that scaled reasoning can translate into appropriate restraint.
The \textit{Gemini} family is the strongest hybrid reasoner, yet remains vulnerable when all required capabilities are internal (only 38.83 KAS).
The \textit{Qwen} and \textit{DeepSeek} families share a sharper profile: their reasoning variants dominate \textit{{External Function}} (97.42 and 97.82 KAS) yet collapse to around 22 in \textit{{Internal Function}}, reflecting a persistent tool-eagerness bias.
No family dominates all three settings, indicating that executing tools and refraining from them are partially independent capabilities rather than two facets of a single skill.

\begin{figure}[t]
    \centering
\includegraphics[width=1\linewidth,trim={0 0 0 0},clip]{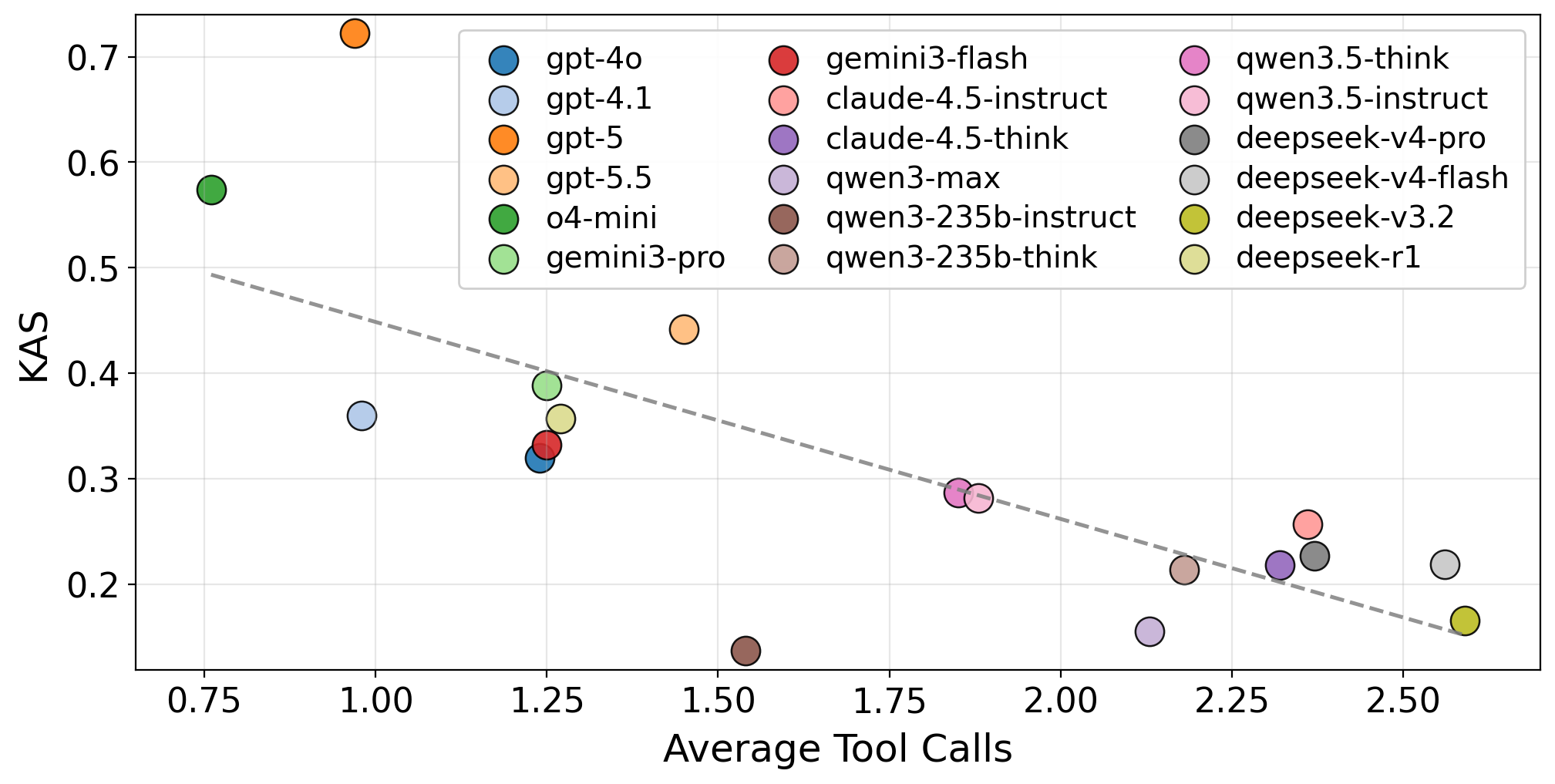}
    \caption{The negative correlation between KAS and tool usage in \textit{{Internal Function}}.}
    \label{fig:tool_calls}
\end{figure}

\section{Further Analysis}
To dissect the underlying  mechanisms, we expand our evaluation from overall performance to analyzing internal behavioral dynamics.

\subsection{RQ1: What are the dominant patterns of mismatch?}

To analyze this interplay, we categorize model outputs as: 
\textit{Consistent Competence} (Kc/Ac), \textit{Alignment Issue} (Kc/Aw), \textit{Behavioral Competence} (Kw/Ac), and \textit{Cognitive Issue} (Kw/Aw), and summarize the dominant side via directional  gap \textit{DirGap} $=p_{Kc/Aw}-p_{Kw/Ac}$: a positive value indicates that Knowing outpaces Acting, while a negative value indicates the opposite. Table~\ref{table:Q1Q2Q3Q4} reveals three setting-specific mechanisms.

\begin{figure*}[h]
\centering
    \begin{subfigure}[b]{0.32\linewidth}
    \centering
      \includegraphics[width=\linewidth]{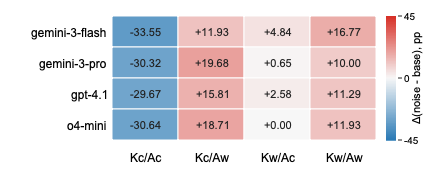}
      \subcaption{\textit{\textbf{External Function.}} Knowing survives but acting fails: $Kc/Ac$ mass flows uniformly into $Kc/Aw$ and $Kw/Aw$.}
    \end{subfigure}
    \begin{subfigure}[b]{0.32\linewidth}
    \centering
    \includegraphics[width=\linewidth]{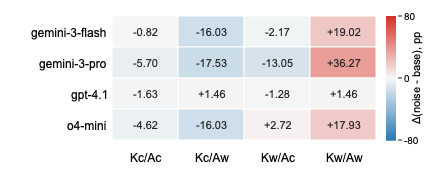}
      \subcaption{\textit{\textbf{Hybrid Composition.}} Errors deepen: $Kc/Aw$ collapses into $Kw/Aw$, marking joint cognitive--behavioral failure.}
    \end{subfigure}
    \begin{subfigure}[b]{0.32\linewidth}
    \centering
      \includegraphics[width=\linewidth]{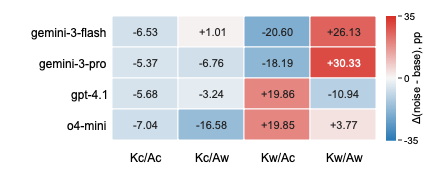}
      \subcaption{\textit{\textbf{Internal Function.}} Trends diverge: $Kw/Ac$ shifts in opposite directions across model families.}
    \end{subfigure}

\caption{Distributional shifts ($\Delta$, in percentage points) of the knowing--acting joint distribution under noisy tool environments, relative to the clean baseline. Each cell reports $\mathrm{P}_{\text{noise}}(\cdot)-\mathrm{P}_{\text{base}}(\cdot)$ over the four quadrants \textit{Consistent Competence} ($Kc/Ac$), \textit{Alignment Issue} ($Kc/Aw$), \textit{Behavioral Competence} ($Kw/Ac$), and \textit{Cognitive Issue} ($Kw/Aw$). Red denotes an increase, blue a decrease.}
\label{fig:heatmap}
\end{figure*}

\paragraph{\textit{\textbf{External Function}}: Knowing precedes acting, and reasoning closes the gap.}
When every task strictly requires external tools, the main failure mode is under-invocation, and a positive DirGap directly localizes the bottleneck on the acting side. 
Further, within most families, reasoning variants markedly narrow this gap, e.g., DirGap drops from $+$42.58 in qwen3-235b-instruct to $+$1.96 in qwen3-235b-think, showing that the thinking stage primarily converts already-correct tool-necessity judgments into the corresponding invocation.

\paragraph{\textit{\textbf{Hybrid Composition}}: Reasoning changes the dominant error type rather than eliminating it.}
Instruct models mainly collapse into the \textit{Cognitive Issue} quadrant,
unable to identify which subtasks demand tools. 
While reasoning sharply reduces this joint failure, the error shifts from Kw/Aw to Kc/Aw: gpt-5.5 reduces Kw/Aw to $27.72\%$ while raising Kc/Aw to $57.34\%$. 
This indicates that reasoning turns a cognition gap into an alignment gap, confirming that recognizing tool necessity is easier than executing the corresponding tool calls.

\paragraph{\textit{\textbf{Internal Function}}: Reasoning sharpens awareness but cannot suppress invocation.}
When the correct action reverses to withholding tools, Kc/Aw now directly captures tool overuse. 
Several instruct models show negative DirGap, meaning they refrain from calling external tools while still misjudging the boundary. 
Reasoning models instead show positive DirGap (claude-sonnet-4.5-think $+$33.67): they correctly recognize that no tool is needed, yet still invoke. 
Reasoning therefore continues to improve \textit{knowing} as in the other two settings, but Acting still fails to refrain from unnecessary tool calls, a failure that reasoning alone does not resolve.

\subsection{RQ2: How does inconsistency impact efficiency?}
To quantify the cost implications of knowing-acting inconsistency, we leverage the \textit{\textbf{Internal Function}} setting as a natural, clean, and strict evaluation environment: since none of these tasks genuinely require external tools, any tool invocation constitutes pure overhead. 
As shown in Figure~\ref{fig:tool_calls}, KAS exhibits a strong negative correlation with the average number of tool calls (Pearson $r{=}-0.748$, $p{<}0.001$). It illustrates that high-consistency models (e.g., gpt-5, o4-mini) use tools at minimal cost ($<1.0$ calls), whereas inconsistent models (e.g., claude-4.5-instruct) incur substantially higher costs ($>2.0$ calls), reflecting redundant and error-prone tool-call loops driven by miscalibrated self-awareness. 
This demonstrates that knowing-acting inconsistency translates directly into efficiency degradation.

\subsection{RQ3: How does reasoning complexity affect the gap?}
We visualized representative models (Figure~\ref{fig:logic_types}) while presenting the complete results in Appendix~\ref{appendix:Additional Experiments}. We find that KAS exhibits a systematic decline trend with increasing logical combinatorial complexity. This indicates  diminished ability of LLM agents to maintain capability self-awareness and consistency under deeper reasoning chains.

\subsection{RQ4: How do noisy tools distort consistency?}
To examine how distractor tools affect knowing--acting consistency, we add noise tools to the candidate list (details in Appendix~\ref{appendix:experiment}) and compare quadrant shifts in Figure~\ref{fig:heatmap}.
Noise reduces consistent success ($Kc/Ac$) in all settings, but the errors move differently.
In \textit{\textbf{External Function}}, the drop in $Kc/Ac$ is split between $Kc/Aw$ and $Kw/Aw$. Thus, models often know tools are needed, but choose or execute the wrong tool.
In \textit{\textbf{Hybrid Composition}}, $Kc/Aw$ drops sharply (${-}16$ to ${-}18$pp) and shifts mainly to $Kw/Aw$, showing that noise hurts both subtask-level tool judgment and action.
In \textit{\textbf{Internal Function}}, the contrast between model family shows that noise can break consistency through different bottlenecks: it exposes weak acting in \textit{Gemini} family, and mainly harms knowing (unable to  recognize whether and which tools are needed)  in \textit{GPT} family.

\section{Conclusion}
This paper introduces the {\textit{\textbf{KAP}\scalebox{0.8}{\textbf{RO}}}} benchmark, designed to evaluate an LLM agent's self-awareness capability by rigorously separating its cognitive self-assessment from behavioral execution. By partitioning tasks into external, hybrid, and internal subspaces with {\textit{\textbf{KA}\scalebox{0.8}{\textbf{ware}}}} dataset, we uncover a critical boundary ambiguity error: agents frequently fail to distinguish between problems solvable through internal knowledge and those requiring external tools. Our extensive evaluation across 18 LLMs demonstrates that current LLM agents possess limited self-awareness, highlighting a structural vulnerability in existing systems. 
We believe this work provides a rigorous diagnostic framework and actionable insights for future LLM agents.
\section{Limitations}
We acknowledge three limitations of this work. First, the parametric--tool boundary  is intrinsically time-dependent, next-generation models with broader pretraining corpora, longer context, or built-in multimodal abilities may internalize tasks we currently label as \emph{External}. Therefore, we plan to continuously update {\textit{\textbf{KA}\scalebox{0.8}{\textbf{ware}}}} to keep its boundary annotations aligned with frontier model capabilities. 
Second, tool discrimination, task synthesis, trajectory acquisition, and final pass-rate judging are all driven by strong LLMs (\textit{gpt-5} for generation and execution, \textit{gpt-4.1} for quality control and judging). Although we mitigate this through rubrics, multi-stage automatic checks, and human re-inspection, residual model-specific biases may still propagate into the values we report for \textit{gpt-5}  and \textit{gpt-4.1}. 
Third, {\textit{\textbf{KAP}\scalebox{0.8}{\textbf{RO}}}} is positioned as a \emph{diagnostic} framework that isolates whether failures originate from a cognitive gap or an executive-control gap, but does not itself propose training-time or inference-time interventions to close these gaps.

\bibliography{custom}

\clearpage

\appendix

\section{Use of AI Assistants}
We used AI assistants to support language polishing. All scientific claims, experimental results, analyses, and final manuscript content were reviewed and verified by the authors.

\section{Additional Related Work}
\label{appendix:relatedworks}

\paragraph{Agent Tool-Use Benchmarks.}
The rapid shift from \emph{chat-only} large language models (LLMs) to \emph{tool-using} agents has motivated a growing set of benchmarks that evaluate whether models can compose multi-step tool calls~\cite{li-etal-2023-api,qin2024toollm,chen-etal-2024-eval,patil2025bfcl,yao2025taubench,mialon2024gaia,lu-etal-2025-toolsandbox,shen-etal-2024-taskbench,yue2025synergistic,guo2025dr,jia2025ready}.
Early benchmarks such as API-Bank and APIBench provide runnable tool environments and annotate tool-use dialogues with API-call traces, emphasizing planning and API retrieval in controlled settings~\cite{li-etal-2023-api, patil2024gorilla, huang-etal-2024-planning-creation}.
ToolBench, introduced with ToolLLM, scales this direction by curating thousands of real-world APIs and automatically constructing tool-use instructions with solution paths, enabling evaluation of tool selection and capability in large tool spaces~\cite{qin2024toollm,guo-etal-2024-stabletoolbench}.
More recent benchmarks further stress compositional tool use. For example, ToolHop proposes a query-driven benchmark for multi-hop tool use, where models must decompose complex queries, identify intermediate dependencies, and invoke multiple tools in a coherent sequence~\cite{ye-etal-2025-toolhop}.
Such benchmarks move beyond single-call correctness and expose failures in long-horizon planning, tool-result integration, and multi-hop reasoning.

Complementary to dataset-centric benchmarks, T-Eval decomposes tool utilization into sub-abilities, such as planning, retrieval, understanding, and review, and evaluates them step-by-step to diagnose where failures occur~\cite{chen-etal-2024-eval,shen-etal-2024-taskbench,zhang-etal-2024-toolbehonest,lu-etal-2025-toolsandbox}.
In parallel, standardized leaderboards such as BFCL focus on function-calling correctness across serial and parallel calls and programming languages via AST-based evaluation, and further include abstention and stateful multi-step settings~\cite{patil2025bfcl,patil2024gorilla}.
Benchmarks like $\tau$-bench move toward realistic agent--user--tool collaboration under stateful environments and outcome-based evaluation~\cite{yao2025taubench,lu-etal-2025-toolsandbox,mialon2024gaia}.
More recently, MCP-oriented evaluations, such as MCP-RADAR and MCPEval, target the emerging Model Context Protocol ecosystem with multi-dimensional metrics or automated end-to-end task generation and verification~\cite{gao2025mcpradar,liu-etal-2025-mcpeval}.
Despite this diversity, existing benchmarks predominantly reward \emph{successful tool invocation} and are largely composed of tool-essential or tool-favored tasks; they provide limited supervision for the \emph{decision to refrain} from tool use when a direct answer is preferable~\cite{patil2025bfcl,zhang-etal-2024-toolbehonest,lu-etal-2025-toolsandbox}.
Our work complements prior benchmarks by explicitly formalizing and annotating the \emph{boundary} between internal capability and external tool need.

\paragraph{LLM Agent Knowledge Boundary.}
Deciding whether to \emph{act} by invoking tools or to \emph{answer} from parametric knowledge is closely related to an agent's knowledge boundary and uncertainty awareness~\cite{kadavath2022lmknow,lin2022uncertaintywords}.
Classical selective prediction with a reject option formalizes the trade-off between coverage and risk, providing a principled view of when models should abstain~\cite{geifman2017selective,chow1970reject,hendrickx2024rejectsurvey}.
Recent evidence suggests that language models can partially estimate what they know: models can learn calibrated self-evaluation signals such as $P(\mathrm{True})$ or $P(\mathrm{IK})$ that correlate with answer correctness~\cite{kadavath2022lmknow,lin2022uncertaintywords}.
On the action side, an agent must translate such self-knowledge into \emph{selective} tool invocation rather than indiscriminate calls. Self-supervised paradigms such as Toolformer train models to learn when and how to call tools~\cite{schick2023toolformer}, and retrieval-aware methods such as Self-RAG further demonstrate that triggering external resources should be conditional on need rather than reflexive~\cite{asai2023selfrag,lewis2020rag,guu2020realm}.

A growing line of recent work studies tool-use gating and tool overuse more directly.
MetaTool explicitly evaluates \emph{whether} to use tools and \emph{which} tools to use via tool-usage awareness and tool-selection tasks~\cite{huang2023metatool}. 
From a theoretical perspective, recent work frames agents as tool-use decision-makers and argues that an agent's tool-use decision boundary should be aligned with its knowledge boundary to avoid unnecessary actions~\cite{wang2025toward}.
Latest further proposes a framework for assessing and optimizing the binary decision of whether to invoke tools, emphasizing that the ability to refrain from tool use is as important as the ability to call tools correctly~\cite{wu2026callframeworkassessoptimize}, and others provides a systematic analysis of why LLMs prefer external tools over internal knowledge~\cite{zeng2026tooloveruseillusiondoesllm}.
Together, these studies suggest that tool use is not merely a capability problem but also a calibration and decision-boundary problem.

However, recent studies primarily target the \emph{tool overuse} regime: redundant invocations when parametric knowledge already suffices, and they rarely release a dedicated boundary-aware evaluation protocol.
We close this gap by explicitly partitioning the task space along the intersection of the \emph{knowledge boundary} and the \emph{tool-use boundary} into three regions: \textit{\textbf{External Function}}, where tools are strictly required; \textit{\textbf{Internal Function}}, where parametric knowledge already suffices; and \textit{\textbf{Hybrid Composition}}, where the two regimes co-occur within a single trajectory.
This partition turns a single benchmark into a \emph{unified} probe of both miscalibration directions: \textit{External Function} isolates \emph{tool underuse}, \textit{Internal Function} isolates \emph{tool overuse}, and \textit{Hybrid Composition} additionally probes sub-step-level joint decision-making, so that overuse and underuse are measured under matched task conditions.

\section{Benchmark Construction}
We use existing models and APIs only for evaluation purposes, following their documented terms of use. 
We will release the benchmark artifacts, including task instances, tool metadata, prompts, and evaluation scripts, under the MIT License. For any third-party resources used in constructing the benchmark, we follow their original licenses and terms of use.
\subsection{Details of Tool Seed Construction}
\label{appendix:tool evaluation}
In this section, we design a multi-stage prompt-based evaluation pipeline that assesses tool functionality, documentation quality, and applicability to reasoning tasks. The following presents the detailed prompts used for tool selection and filtering.

\begin{tcolorbox}[
    enhanced,
    breakable,
    colback=violet!3,       
    colframe=violet!30!black, 
    colbacktitle=violet!30!black, 
    coltitle=white,
    boxrule=2pt,
    arc=0mm,
    left=10pt,
    right=10pt,
    top=10pt,
    bottom=10pt,
    fonttitle=\bfseries\large,
    title={Tool Evaluation Prompt},
    attach boxed title to top left={yshift=-2mm}
]

You are a strict agent tool analyst. Your job is to evaluate an external tool against the boundaries of a Large Language Model, and filter it into high-confidence categories for a tool library.

\vspace{0.5em}

\textbf{Crucial Rule:} You must be conservative. If the tool description is vague, missing parameters, or you are not confident about a dimension, you \textbf{MUST} flag that dimension as \textbf{``Unknown''}.

\vspace{0.5em}

\textbf{Tool:} \{tool\_description\}\\
\textbf{OUTPUT:} JSON only\\
\textbf{STATUS OPTIONS:} True / False / Unknown

\vspace{0.5em}
\hrule
\vspace{0.5em}

\subsection*{EVALUATION FRAMEWORK:}
Analyze the tool across the following 6 dimensions. Use these status definitions:

\begin{itemize}[leftmargin=*, topsep=0pt, itemsep=0pt, parsep=0pt]
    \item \textbf{True}: There is \textbf{CLEAR evidence} the tool hits a hard constraint.
    \item \textbf{False}: There is \textbf{CLEAR evidence} the tool is safe/native.
    \item \textbf{Unknown}: The description is vague or you cannot determine the truth.
\end{itemize}

\subsubsection*{1. Interaction Scope}
\textbf{Assessment Logic:} Does the tool alter the state of the world or system?
\begin{itemize}[leftmargin=*, topsep=0pt, itemsep=0pt, parsep=0pt]
    \item \textbf{True:} State-mutating operations (e.g., Create/Update/Delete, ``Send Email'', ``Buy Stock'', ``Reboot Server'').
    \item \textbf{False:} Read-only or informational (e.g., ``Get Weather'', ``Search Docs'', ``Read File'').
    \item \textbf{Unknown:} Vague verbs without specifying read/write.
\end{itemize}

\subsubsection*{2. Data Privacy Scope}
    \textbf{Assessment Logic:} Does the tool require restricted data access?
\begin{itemize}[leftmargin=*, topsep=0pt, itemsep=0pt, parsep=0pt]

    \item \textbf{True:} Accesses private/sensitive data (e.g., PII, internal DBs, user-auth content).
    \item \textbf{False:} Uses public-domain knowledge (e.g., Wikipedia facts, general reasoning, open web).
    \item \textbf{Unknown:} Data source/permission level not mentioned.
\end{itemize}

\subsubsection*{3. Data Temporal Scope}
    \textbf{Assessment Logic:} Does the tool require real-time synchronization?
\begin{itemize}[leftmargin=*, topsep=0pt, itemsep=0pt, parsep=0pt]

    \item \textbf{True:} Depends on real-time/dynamic data (e.g., current stock price, live traffic, breaking news).
    \item \textbf{False:} Uses static/historical knowledge or timeless principles.
    \item \textbf{Unknown:} Time sensitivity unclear.
\end{itemize}

\subsubsection*{4. Modality Scope}
    \textbf{Assessment Logic:} Does the tool handle data formats or media types unsupported by standard LLMs?
\begin{itemize}[leftmargin=*, topsep=0pt, itemsep=0pt, parsep=0pt]

    \item \textbf{True:} Unsupported/specialized modalities (e.g., video generation, 3D models/point clouds, CAD files).
    \item \textbf{False:} Standard native modalities (e.g., text, code).
    \item \textbf{Unknown:} Input/output media format not specified.
\end{itemize}

\subsubsection*{5. Computational Scope}
     \textbf{Assessment Logic:} Does the task require rigorous calculation or exact execution?
\begin{itemize}[leftmargin=*, topsep=0pt, itemsep=0pt, parsep=0pt]

    \item \textbf{True:} Hard logic where LLMs often fail (e.g., complex math, cryptography, code execution).
    \item \textbf{False:} Semantic reasoning or basic arithmetic (e.g., summarization, classification, creative writing, \texttt{1+1}).
    \item \textbf{Unknown:} Computational complexity unclear.
\end{itemize}

\subsubsection*{6. Data Scale Scope}
    \textbf{Assessment Logic:} Does the data volume exceed context limits?
\begin{itemize}[leftmargin=*, topsep=0pt, itemsep=0pt, parsep=0pt]

    \item \textbf{True:} Out-of-core processing (e.g., terabytes of logs, bulk DB aggregation).
    \item \textbf{False:} In-context processing (fits within a standard prompt window).
    \item \textbf{Unknown:} Data volume not specified.
\end{itemize}

\vspace{0.5em}
\hrule
\vspace{0.5em}

\subsection*{FINAL CLASSIFICATION RULES:}
Apply these rules in order to determine \texttt{final\_classification}:
\begin{itemize}[leftmargin=*, topsep=0pt, itemsep=0pt, parsep=0pt]
    \item \textbf{True}: If \textbf{ANY} dimension is \texttt{True}.
    \item \textbf{False}: If and only if \textbf{ALL} dimensions are \texttt{False}.
    \item \textbf{Unknown}: If no dimension is \texttt{True}, and at least one dimension is \texttt{Unknown}.
\end{itemize}

\vspace{0.5em}
\hrule
\vspace{0.5em}

\subsection*{Required JSON Output Format (STRICT):}
\begin{lstlisting}[%
    basicstyle=\ttfamily\footnotesize,  % 小字体
    breaklines=true,                    % 自动换行
    breakatwhitespace=true           % 只在空格换行
]
{
  "tool_name": "{Tool Name}",
  "dimensions": {
    "interaction_scope": {
      "status": "True" | "False" | "Unknown",
      "reasoning": "Brief explanation.",
      "evidence": "Quote specific content."
    },
    "data_privacy_scope": {
      "status": "True" | "False" | "Unknown",
      "reasoning": "Brief explanation.",
      "evidence": "Quote specific content."
    },
    "data_temporal_scope": {
      "status": "True" | "False" | "Unknown",
      "reasoning": "Brief explanation.",
      "evidence": "Quote specific content."
    },
    "modality_scope": {
      "status": "True" | "False" | "Unknown",
      "reasoning": "Brief explanation.",
      "evidence": "Quote specific content."
    },
    "computational_scope": {
      "status": "True" | "False" | "Unknown",
      "reasoning": "Brief explanation.",
      "evidence": "Quote specific content."
    },
    "data_scale_scope": {
      "status": "True" | "False" | "Unknown",
      "reasoning": "Brief explanation.",
      "evidence": "Quote specific content."
    }
  },
  "final_classification": "True" | "False" | "Unknown"
}
\end{lstlisting}

\end{tcolorbox}

\subsubsection{Tool Statistics}
\paragraph{Annotation Pipeline Statistics.}
The pipeline partitions every candidate tool into one of two coarse categories: \textit{Definitely Outside} the parametric capability boundary: tool operations that demonstrably require external execution (e.g., live retrieval, side effects, multimodal I/O), and \textit{Potentially Inside} the boundary: tool operations whose functionality could be subsumed by the model's internal knowledge. This binary partition is motivated by two complementary considerations. First, it yields a curated tool seed pool on which the downstream task synthesis can be reliably anchored, avoiding semantically ill-defined or boundary-ambiguous operations. Second, by restricting the subsequent sampling space to tools with clearly delineated boundary status, we expect the boundary-aware task filtering to retain a higher fraction of generated candidates. 

Considering exhaustively annotating all  deduplicated entries (65{,}770) with a single high-capacity model would be prohibitively expensive, we adopt a two-stage cascade that explicitly trades coverage in Stage~1 for precision in Stage~2: an open-source screener (Qwen3-30B-A3B) first performs high-recall coarse labeling, and a stronger reviewer (GPT-4.1) is then invoked to refine the labels; a tool's final label is committed only when the two stages agree. A held-out human-annotated subset further audits the reliability of the automated labels. Table~\ref{tab:annotation_stats} reports the per-stage counts, the cross-model agreement rate, and the human-verification statistics.

\paragraph{Human Verification Protocol.}
To audit the reliability of the automated cascade, we recruit three PhD-level annotators with prior research experience in LLM agents and tool-use systems. We draw a validation subset $\mathcal{T}_{\text{val}} \subset \mathcal{T}$ of 400 tools from the post-cascade pool, stratified by the three Stage~2 labels (\textit{Definitely Outside}, \textit{Potentially Inside}, and \textit{Unknown}) so that the minority classes are adequately represented. To reduce potential bias, all annotators received identical task instructions and the same six-dimensional rubric used by the automated screener: \textit{Interaction Scope}, \textit{Temporality Scope}, \textit{Data Privacy Scope}, \textit{Data Scale Scope}, \textit{Computational Scope}, and \textit{Modality Scope}, and completed the annotation independently. The automated labels and the identities of both screening and reviewing models were concealed throughout the process, so that every judgment is grounded solely in the standardized tool schema $\mathcal{S}_{\text{tool}} = \langle \text{name, params, return, desc} \rangle$. To mitigate the impact of individual outliers, the final human label of each tool is aggregated by majority vote over the three annotators; the rare three-way disagreements are resolved through a post-hoc consensus discussion among the annotators. We then compare the aggregated human labels against the committed automated labels, obtaining an overall agreement of 90.41\% on $\mathcal{T}_{\text{val}}$ (Table~\ref{tab:annotation_stats}), which empirically supports the reliability of the cascade.

\begin{table*}[t]
\centering
\setlength{\tabcolsep}{6pt}
\renewcommand{\arraystretch}{1.08}
\caption{Statistics of the cascaded annotation pipeline. \textit{Stage 1} represents the initial screening by Qwen3-30B-A3B, and \textit{Stage 2} represents the refinement by GPT-4.1. The \textit{Consensus} section reflects samples where both models aligned.}
\label{tab:annotation_stats}
\begin{tabular}{@{}llr@{}}
\toprule
\textbf{Pipeline Phase} & \textbf{Metric/Category} & \textbf{Count/Value} \\
\midrule
\multirow{3}{*}{\textbf{Stage 1}}
 & Definitely Outside & 32,994 \\
 & Potentially Inside & 2,515 \\
 & Unknown & 30,261 \\
\midrule
\multirow{3}{*}{\textbf{Stage 2}}
 & Definitely Outside & 29,905 \\
 & Potentially Inside & 1,731 \\
 & Unknown & 3,873 \\
\midrule
\multirow{3}{*}{\textbf{Consensus}}
 & Consistent Samples & 28,859 \\
 & Inconsistent Samples & 4,112 \\
 & Agreement Rate & 87.53\% \\
\midrule
\multirow{2}{*}{\textbf{Human Verification}}
 & Sample Size & 400 \\
 & Human Agreement & 90.41\% \\
\bottomrule
\end{tabular}
\end{table*}

\begin{figure}[H]
    \centering
    \includegraphics[width=1.0\linewidth]{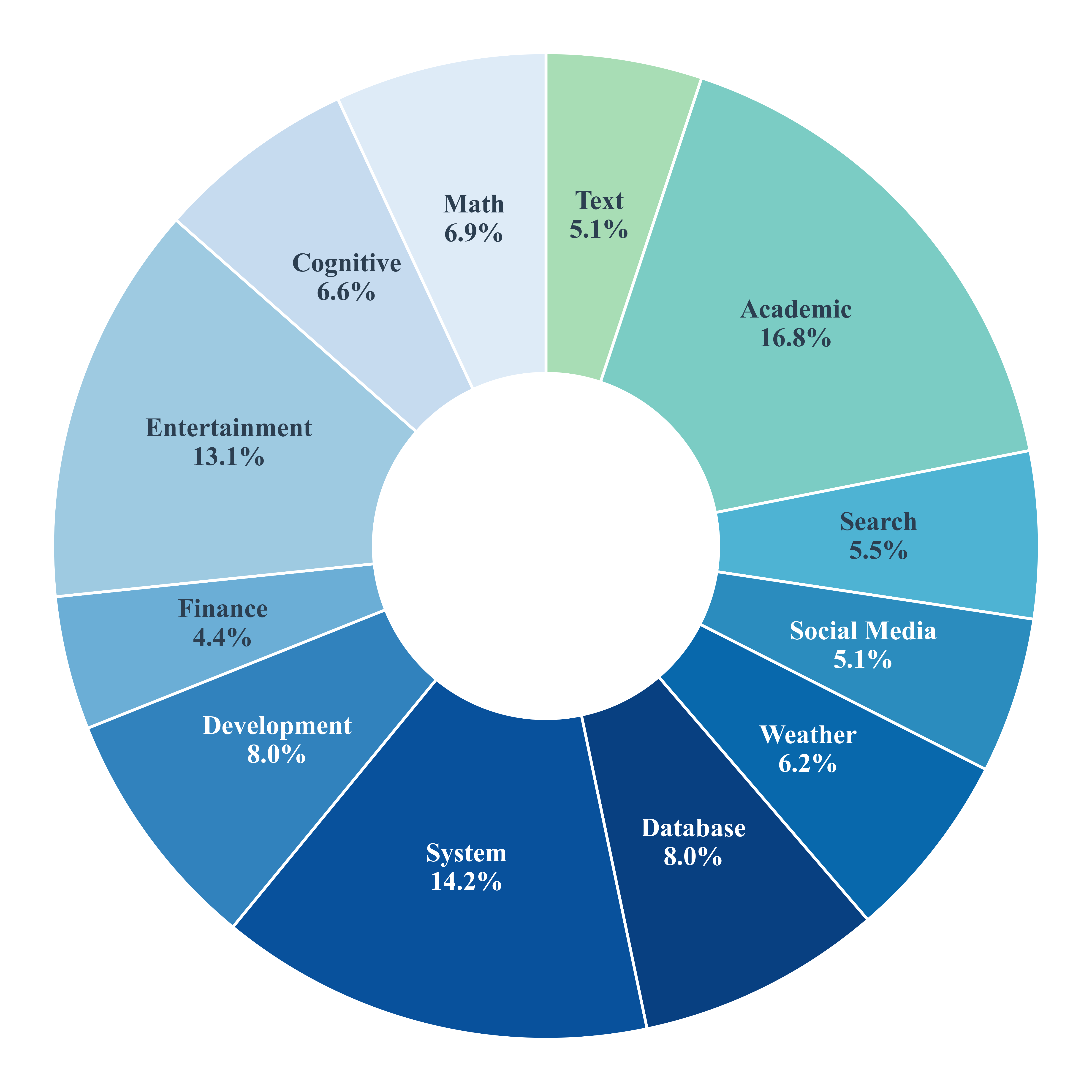}
    \caption{Tool distribution by domain, with parentheses showing the percentage of tools in each category.}
    \label{fig:tool_stats}
\end{figure}

We further present statistics of the final selected tool set. Figure~\ref{fig:tool_stats} visualizes the tool distribution across different domains.
\label{appendix:tool statistics}

\subsection{Details of Task Generation and Quality Control}
\label{appendix:task gen and quality}
In this section, we employ a two-stage pipeline combining task generation prompts with rigorous quality filtering. Human inspection results in Table~\ref{tab:accuracy-by-type} confirm high annotation agreement across all logic types, validating our data construction process.

\begin{tcolorbox}[
    enhanced,
    breakable,
    colback=violet!3,              
    colframe=violet!30!black,      
    colbacktitle=violet!30!black,  
    coltitle=white,                
    boxrule=2pt,
    arc=0mm,
    left=10pt,
    right=10pt,
    top=10pt,
    bottom=10pt,
    fonttitle=\bfseries\large,
    title={Tool Use Question Generation Prompt}, 
    attach boxed title to top left={yshift=-2mm}
]

\subsection*{Task}
Generate a high-quality \textbf{Tool Use Question} based on the provided MCP Server and its tool descriptions.

\subsection*{Objective}
Analyze the provided MCP Server and its available tools to create a realistic user request. The request must strictly require the use of \textbf{\{NUM\_TOOLS\} tools} and must fall into one of two specific logical patterns: \textbf{Parallel (Decomposed)} or \textbf{Multi-hop (Sequential)}.

\subsection*{Guidelines}

\subsubsection*{Dependency \& Logic Analysis}
You must analyze the relationship between the chosen \{NUM\_TOOLS\} tools and define the data flow:
\begin{itemize}
    \item \textbf{PARALLEL:}
    \begin{itemize}
        \item The tools function independently.
        \item Used for comparing data (e.g., "Price of A vs Price of B") or aggregating different aspects (e.g., "Weather and Traffic").
        \item \textit{Constraint:} No tool output is required as input for another tool.
    \end{itemize}
    \item \textbf{MULTI-HOP:}
    \begin{itemize}
        \item Strict input-output dependency exists.
        \item \textit{Constraint:} You MUST identify the specific field from Tool A's output that feeds into Tool B's input (e.g., \texttt{Tool A output[id] -> Tool B input[user\_id]}).
    \end{itemize}
\end{itemize}

\subsubsection*{Knowledge Boundary Enforcement}
\begin{itemize}
    \item The question MUST require external information or action that cannot be answered from the LLM's training data.
    \item \textbf{Do not} ask general knowledge questions (e.g., "What is Python?").
    \item \textbf{Do not} ask questions that can be answered by reasoning alone.
    \item Focus on one of these categories:
    \begin{itemize}
        \item \textbf{Real-time Data:} Stock prices, current weather, system status, recent logs.
        \item \textbf{Private Data:} A specific user's database record, a local file content, a private repository issue.
        \item \textbf{Actions:} Sending an email, rebooting a server, creating a file, toggling a smart switch.
    \end{itemize}
    \item The combination of \textbf{\{NUM\_TOOLS\} tools} should work together to access information or perform actions that are external to the LLM's knowledge base.
\end{itemize}

\subsubsection*{Realistic Scenario and Persona}
\begin{itemize}
    \item Adopt a specific persona relevant to the tools (e.g., a DevOps Engineer checking server health, a Sales Manager looking up a CRM record).
    \item The question should sound like a task, not a test.
    \item \textbf{Example of bad quality:} "Use the tools to find the weather and send an email."
    \item \textbf{Example of good quality (sub-questions format):} "What's the weather forecast for Yosemite Valley tomorrow morning? Can you send a reminder email to the team with the weather forecast?"
\end{itemize}

\subsubsection*{Temporal Constraint}
\begin{itemize}
    \item \textbf{Mandatory Date Shift:} Assume the current date is \textbf{February 2025 or later}, or ensures the task targets a date/deadline in this future range.
    \item If the tools involve scheduling, checking availability, or retrieving time-sensitive data (stock prices, logs, weather), the context must reflect this timeframe (e.g., "June 16th, 2025").
\end{itemize}

\subsubsection*{Question Complexity}
\begin{itemize}
    \item Create questions that are clear and specific enough to warrant using \{NUM\_TOOLS\} tools.
    \item The question should have multiple components or require several steps to solve.
    \item Include relevant context or constraints that make the multi-tool usage necessary.
    \item Do not contain the exact tool names in the question.
    \item \textbf{Sub-Questions (Atomic):} Clear, distinct steps separated by \texttt{?}, each corresponding to one tool in the \texttt{target\_tools} list, and listed in the same order as the tools in \texttt{target\_tools}.
    \item \textbf{Natural Task (Conversational):} Describe the \textbf{User's Intent}, not the technical steps.
\end{itemize}

\subsection*{Output Format}
Provide your response in a valid \textbf{XML} object with the following structure:
\begin{enumerate}
    \item \texttt{server\_analysis}: A brief analysis of the MCP Server's domain and capabilities.
    \item \texttt{dependency\_analysis}: An object containing:
    \begin{itemize}
        \item \texttt{logic\_type}: Strictly either \texttt{"PARALLEL"} (independent tools) or \texttt{"multi-hop"} (sequential dependency).
        \item \texttt{data\_flow}: Describe the parameter passing.
        \item For \textbf{Multi-hop}: Format as \texttt{"Tool A (output\_field) -> Tool B (input\_field)"}.
        \item For \textbf{Parallel}: Set to \texttt{"None"}.
        \item \texttt{reasoning}: A brief explanation of why this sequence requires the chosen logic type.
    \end{itemize}
    \item \texttt{target\_tools}: A list of exactly \textbf{\{NUM\_TOOLS\}} tool names from the MCP Server, listed in the order they must be called.
    \item \texttt{sub\_questions}: A string containing \textbf{\{NUM\_TOOLS\}} distinct atomic questions or actions, separated by \texttt{?}. Each segment must correspond exactly to one tool in \texttt{target\_tools} in the same order.
    \item \texttt{natural\_task}: A single, coherent, natural language user request that combines the sub-questions into a realistic scenario. This should focus on the user's intent and avoid mentioning technical tool names or robotic steps.
\end{enumerate}

\subsection*{MCP Server Description}
\{MCP\_SERVER\_NAME\}: \{MCP\_SERVER\_DESCRIPTION\}

Available Tools:
\{TOOL\_LIST\}

\subsection*{Output}
Ensure your question requires exactly \{NUM\_TOOLS\} tools to solve completely. Provide your response in the following XML format:

\begin{verbatim}
<response>
  <server_analysis>
    </server_analysis>
  <dependency_analysis>
    <logic_type>
      </logic_type>
    <data_flow>
      </data_flow>
    <reasoning>
      </reasoning>
  </dependency_analysis>
  <target_tools>
    <tool>tool_name</tool>
  </target_tools>
  <sub_questions>
    </sub_questions>
  <natural_task>
    </natural_task>
</response>
\end{verbatim}

\end{tcolorbox}
\begin{tcolorbox}[
    enhanced,
    breakable,
    colback=violet!3,
    colframe=violet!40!black,
    colbacktitle=violet!40!black,
    coltitle=white,
    boxrule=1.5pt,
    arc=1mm,
    left=12pt,
    right=12pt,
    top=10pt,
    bottom=10pt,
    fonttitle=\bfseries\large,
    title={Tool Scenario Prompt},
    attach boxed title to top left={yshift=-2mm, xshift=2mm}
]

\textbf{Role}\par\smallskip
You are a \textbf{Technical Use-Case Architect}.
Your objective is to translate technical tool definitions into compelling, realistic, and practical real-world narratives.

\medskip
\textbf{Input Data}\par\smallskip
\texttt{\{tool\_name\}} \\
\texttt{\{tool\_description\}} \\
\texttt{\{input\_schema\}}

\medskip
\textbf{Task Description}\par\smallskip
Generate a concise, realistic application scenario based on the provided MCP tool information.
The scenario must bridge the gap between technical capability and user value.

\medskip
\textbf{Generation Requirements}\par\smallskip
\begin{itemize}
    \item \textbf{Contextual Clarity}: Clearly describe a specific real-world situation or problem (e.g., debugging, data analysis, automation).
    \item \textbf{Practical Application}: Explain explicitly how the tool is used to solve that problem.
    \item \textbf{Schema Alignment}: The scenario must be logically consistent with the provided input schema (parameters/arguments).
    \item \textbf{Conciseness}: Strictly adhere to the 2--4 sentence limit.
    \item \textbf{Language}: Standard Professional English.
\end{itemize}

\medskip
\textbf{Structure of the Scenario}\par\smallskip
The output should seamlessly cover these three elements:
\begin{enumerate}
    \item \textbf{The Trigger}: Who is the user and what is their immediate pain point?
    \item \textbf{The Action}: How is this specific tool invoked to handle the situation?
    \item \textbf{The Outcome}: What is the immediate value or result gained?
\end{enumerate}

\medskip
\textbf{Output Format}\par\smallskip
Return \textbf{only} the paragraph containing the scenario. Do not include headers, explanations, or introductory text.

\end{tcolorbox}
\begin{tcolorbox}[
    enhanced,
    breakable,
    colback=violet!3,
    colframe=violet!40!black,
    colbacktitle=violet!40!black,
    coltitle=white,
    boxrule=1.5pt,
    arc=1mm,
    left=12pt,
    right=12pt,
    top=10pt,
    bottom=10pt,
    fonttitle=\bfseries\large,
    title={Question Quality Assessment Prompt},
    attach boxed title to top left={yshift=-2mm, xshift=2mm}
]

\textbf{Role}\par\smallskip
You are a \textbf{Question Quality Assessment Analyst}.
Your task is to analyze the provided tool use question and assess its quality across six key dimensions based strictly on the provided criteria.

\medskip
\textbf{Task and Input Information}\par\smallskip
\texttt{\{SERVER\_INFORMATION\}} \\
\texttt{\{TOOL\_INFORMATION\}} \\
\texttt{\{QUESTION\_CONTENT\}} \\
\texttt{\{INTENDED\_TOOL\}}

\medskip
\textbf{Evaluation Framework}\par\smallskip
Analyze the question across the following six dimensions.
For each dimension, provide a binary judgment: \textbf{Yes} or \textbf{No}.

\textbf{Judgment Rules:}
\begin{itemize}
    \item \textbf{Yes}: The specific criteria for a positive classification are met.
    \item \textbf{No}: The criteria are not met, or the condition is absent.
\end{itemize}

You must provide detailed reasoning before the rating for each metric.

\medskip
\textbf{1. Is the Question Beyond LLM Knowledge Boundary?}\par\smallskip
\textbf{Definition:} Is it impossible for the LLM to answer correctly using only internal training data?
\begin{itemize}
    \item \textbf{Yes}: Asks for \textbf{private/proprietary data} (e.g., "Check server logs") or \textbf{real-time/future events} (e.g., "What is the weather now?").
    \item \textbf{No}: Asks for general knowledge, coding syntax, or historical facts (e.g., "Capital of France").
\end{itemize}

\medskip
\textbf{2. Is Tool Selection Challenging?}\par\smallskip
\textbf{Definition:} Does the model need to infer the tool from the goal rather than explicit instructions?
\begin{itemize}
    \item \textbf{Yes}: No specific tool name is mentioned; mapping intent requires reasoning.
    \item \textbf{No}: Explicitly names the function/API, or the intent is trivially obvious (1-to-1 match).
\end{itemize}

\medskip
\textbf{3. Is Tool Selection Unique?}\par\smallskip
\textbf{Definition:} Is the selected tool the only logical way to solve the problem?
\begin{itemize}
    \item \textbf{Yes}: No redundant tools exist; specific capabilities are required.
    \item \textbf{No}: Multiple tools could solve the problem equally well.
\end{itemize}

\medskip
\textbf{4. Is Parameter Complete?}\par\smallskip
\textbf{Definition:} Does the question naturally contain the arguments (ID, Time, Location) required for execution?
\begin{itemize}
    \item \textbf{Yes}: Includes specific details (e.g., "Restart \textbf{prod-db-01}").
    \item \textbf{No}: Lacks specific targets, forcing guesses (e.g., "Restart the server").
\end{itemize}

\medskip
\textbf{5. Is Scenario Realistic?}\par\smallskip
\textbf{Definition:} Does this represent a genuine user need in a real-world workflow?
\begin{itemize}
    \item \textbf{Yes}: Sounds like a request from a developer, operator, or real user.
    \item \textbf{No}: The scenario feels contrived, artificial, or theoretically impossible.
\end{itemize}

\medskip
\textbf{6. Is the Answer Verifiable?}\par\smallskip
\textbf{Definition:} Can the correctness of the final answer be objectively proven?
\begin{itemize}
    \item \textbf{Yes}: Relies on objective data (numbers, status codes, specific strings).
    \item \textbf{No}: The expected answer is purely subjective, creative, or open-ended.
\end{itemize}

\medskip
\textbf{Output Format}\par\smallskip
You MUST respond using exactly the following XML format. Do not add any text before or after.

\medskip
\begin{verbatim}
<response>
  <beyond_llm_knowledge>
    <reasoning>
      </reasoning>
    <rating></rating>
  </beyond_llm_knowledge>

  <tool_selection_challenging>
    <reasoning>
      </reasoning>
    <rating></rating>
  </tool_selection_challenging>

  <tool_selection_uniqueness>
    <reasoning>
      </reasoning>
    <rating></rating>
  </tool_selection_uniqueness>

  <parameter_completeness>
    <reasoning>
      </reasoning>
    <rating></rating>
  </parameter_completeness>

  <scenario_realism>
    <reasoning>
      </reasoning>
    <rating></rating>
  </scenario_realism>

  <answer_verifiable>
    <reasoning>
      </reasoning>
    <rating></rating>
  </answer_verifiable>
</response>
\end{verbatim}

\end{tcolorbox}
\begin{tcolorbox}[
    enhanced,
    breakable,
    colback=violet!3,
    colframe=violet!40!black,
    colbacktitle=violet!40!black,
    coltitle=white,
    boxrule=1.5pt,
    arc=1mm,
    left=12pt,
    right=12pt,
    top=10pt,
    bottom=10pt,
    fonttitle=\bfseries\large,
    title={Data Evaluation Prompt},
    attach boxed title to top left={yshift=-2mm, xshift=2mm}
]

\textbf{Role}\par\smallskip
You are a \textbf{Strict Tool-Use and Answer Evaluation Analyst}.  
Your task is to conservatively evaluate tool usage and final answer correctness based only on the provided information.

\medskip
\textbf{Core Rule}\par\smallskip
If evidence is vague, missing, or insufficient, you must return \textbf{False}.  
Do not guess, assume, or rely on external knowledge.

\medskip
\textbf{Task and Tool Information}\par\smallskip
\texttt{\{Sample Information\}} \\
\texttt{\{Ground Truth Information\}} \\
\texttt{\# Information contain task, tool documentation, tool dependency, tool response, answer}

\medskip

\textbf{Label definitions:}
\begin{itemize}
    \item \textbf{True}: There is clear and explicit evidence supporting correctness.
    \item \textbf{False}: There is evidence of violation, misuse, incorrectness, or insufficient information.
\end{itemize}

You must provide a short justification (1--2 sentences) for the decision.

\medskip
\textbf{General Constraints}\par\smallskip
\begin{enumerate}
    \item Rely strictly on tool documentation, task and answer.
    \item Do not use external knowledge or common sense.
\end{enumerate}

\medskip

\textbf{Assessment Logic:}  
Determine whether the final answer accurately completes the task.
The answer must be consistent with the ground truth answer and must not contradict any relevant tool response.

\textbf{Judgment Rules:}
\begin{itemize}
    \item \textbf{True}: The answer is accurate, supported by  the ground truth answer and tool response, and fully addresses the task.
    \item \textbf{False}: The answer is incorrect, hallucinatory, unsupported, inconsistent, or incomplete.
\end{itemize}

\medskip
\textbf{Output Format}\par\smallskip
You MUST respond using exactly the following format.  
Do not add any text before or after.

\medskip
\begin{verbatim}
<True / False>
Reason: <brief explanation>
\end{verbatim}

\end{tcolorbox}

\subsection{Human Evaluation}
\label{appendix: dataset Human Evaluation}
To provide an independent check beyond the automated quality filtering in \S\ref{subsec:synthesis_pipeline}, three PhD-level annotators re-inspect a randomly sampled subset of the constructed benchmark. We draw the audit set stratified by logic type so that each split is adequately represented, producing $141$ instances in total ($49$ from \textit{External Function}, $44$ from \textit{Hybrid Composition}, and $48$ from \textit{Internal Function}; approximately $13\%$ of the full dataset, with per-split ratios reported in Table~\ref{tab:accuracy-by-type}).
For every audited instance, each annotator independently verifies three orthogonal facets: (i) the \textit{task description} is fluent, self-contained, and faithfully reflects the intended user query; (ii) the \textit{logic-type annotation} (single-hop, multi-hop, or parallel) and the \textit{capability-boundary label} (\textit{External}, \textit{Internal}, or \textit{Hybrid}) are correctly assigned; and (iii) the \textit{tool dependency structure}---the dependency graph $G_{\text{dep}}$ together with the ground-truth invocation order---is logically consistent with the task description and the declared boundary. An instance is marked as \emph{agreed} only when all three facets pass; otherwise it is flagged as a failure case for diagnosis.
To reduce potential bias, all annotators received identical task guidelines, completed the inspection independently, and were blinded to both the model identity and the automated quality scores. The per-instance verdict is aggregated by majority vote over the three annotators, and the rare three-way disagreements are resolved through a post-hoc consensus discussion. As reported in Table~\ref{tab:accuracy-by-type}, the human-versus-automated agreement remains high across all three logic types ($\geq 90.91\%$ per split, $96.45\%$ aggregate).
Here, human verification was conducted by three PhD-level student annotators with research experience in NLP and LLM-based tool use. No crowdsourcing platform or external paid participant pool was used.

\begin{table*}[t]
\centering
\small
\setlength{\tabcolsep}{5pt}
\renewcommand{\arraystretch}{1.10}
\caption{Per-split human inspection results on the \our{} benchmark. Three PhD-level annotators independently audit a stratified random subset (covering $\sim\!13\%$ of the dataset). \textbf{Agreement} denotes the fraction of audited instances for which the majority-vote human verdict matches the automated label across all three facets (task description, logic-type annotation, and tool dependency structure).}
\label{tab:accuracy-by-type}
\begin{tabular}{l c c c}
\toprule
\textbf{Logic Type} & \textbf{Audited / Total} & \textbf{Coverage (\%)} & \textbf{Agreement (\%)} \\
\midrule
\textit{External Function}  & 49 / 310     & 15.81 & 100.00 \\
\textit{Hybrid Composition} & 44 / 368     & 11.96 & 90.91  \\
\textit{Internal Function}  & 48 / 398     & 12.06 & 97.92  \\
\midrule
\textbf{Overall}            & 141 / 1{,}076 & 13.10 & 96.45  \\
\bottomrule
\end{tabular}
\end{table*}

\section{Details of Evaluation Metrics}
\label{appendix:metrics}
This appendix establishes two formal benefits of defining KAS as the harmonic mean of two \emph{grounded} Jaccard accuracies:
(i) the underlying Jaccard error decomposes naturally into directional components (tool overuse vs.\ tool underuse), so KAS extends to interpretable error analysis without any extra modeling assumption;
(ii) KAS is strictly stronger than raw know--act agreement, ruling out two failure modes that a pure consistency score cannot detect.

\paragraph{Natural extension to tool overuse and underuse.}
KAS is built from the grounded accuracies $\mathrm{Acc}_{\mathrm{know}}$ and $\mathrm{Acc}_{\mathrm{act}}$, whose complementary Jaccard errors conflate two qualitatively different failures: invoking unnecessary tools and omitting required ones. For predicted set $X$ and reference $T$, define
\begin{align*}
d_J(X,T) &= 1-J(X,T), \\
O(X,T)  &= \frac{|X\setminus T|}{|X\cup T|}, \\
U(X,T)  &= \frac{|T\setminus X|}{|X\cup T|},
\end{align*}
with all three set to $0$ when $X=T=\emptyset$. The symmetric difference $X\triangle T=(X\setminus T)\sqcup (T\setminus X)$ is a disjoint union of subsets of $X\cup T$, so
\begin{equation}
d_J(X,T)=O(X,T)+U(X,T).
\end{equation}
For $s\in\{\mathrm{know},\mathrm{act}\}$, the dataset-level means
\begin{align*}
\mathrm{ToolOver}_{s}  &= \tfrac{1}{|D|}\sum\nolimits_{i=1}^{|D|} O\!\left(E_{s}^{(i)},T^{(i)}\right),\\
\mathrm{ToolUnder}_{s} &= \tfrac{1}{|D|}\sum\nolimits_{i=1}^{|D|} U\!\left(E_{s}^{(i)},T^{(i)}\right)
\end{align*}
inherit the per-instance identity by linearity:
\begin{equation}
1-\mathrm{Acc}_{s}=\mathrm{ToolOver}_{s}+\mathrm{ToolUnder}_{s}.
\end{equation}
Every unit of accuracy loss in KAS is therefore attributable to a specific direction of error on either the knowing or the acting branch.

\paragraph{Stricter than raw know--act agreement.}
A natural alternative to KAS is the raw know--act agreement
\begin{equation}
\overline{C}_{\mathrm{KA}}=\tfrac{1}{|D|}\sum\nolimits_{i=1}^{|D|} J\!\left(E_{\mathrm{know}}^{(i)}, E_{\mathrm{act}}^{(i)}\right),
\end{equation}
which only measures consistency between the two branches. By grounding both branches in $T^{(i)}$ and combining them via the harmonic mean, KAS rules out two failure modes that $\overline{C}_{\mathrm{KA}}$ cannot distinguish.

\smallskip
\noindent\textbf{(1) Jointly wrong tool sets.}
An agent that confidently predicts \emph{and} invokes the same wrong tool set attains $\overline{C}_{\mathrm{KA}}=1$ while failing the task. KAS suppresses this case by lower-bounding $\overline{C}_{\mathrm{KA}}$: high KAS forces high $\overline{C}_{\mathrm{KA}}$, but not conversely. Concretely, the Jaccard distance is a metric on finite sets, so the triangle inequality gives, per task,
\begin{equation}
\begin{split}
d_J\!\left(E_{\mathrm{know}}^{(i)},E_{\mathrm{act}}^{(i)}\right) &\leq d_J\!\left(E_{\mathrm{know}}^{(i)},T^{(i)}\right) \\
&\quad + d_J\!\left(E_{\mathrm{act}}^{(i)},T^{(i)}\right).
\end{split}
\end{equation}
Averaging over $D$ and substituting $d_J=1-J$ yields $\overline{C}_{\mathrm{KA}}\geq a+b-1$, where $a=\mathrm{Acc}_{\mathrm{know}}$ and $b=\mathrm{Acc}_{\mathrm{act}}$. Since $(a-b)^2\geq 0$ implies $(a+b)^2\geq 4ab$, we have $a+b\geq 4ab/(a+b)=2\,\mathrm{KAS}$, and therefore
\begin{equation}
\overline{C}_{\mathrm{KA}}\;\geq\;2\,\mathrm{KAS}-1.
\end{equation}
Once $\mathrm{KAS}>\tfrac{1}{2}$ this certifies high raw agreement; the converse fails, since shared-misconception pairs satisfy $\overline{C}_{\mathrm{KA}}=1$ with $\mathrm{KAS}=0$.

\smallskip
\noindent\textbf{(2) One-sided competence.}
Raw agreement also tolerates a strong branch carrying a weak one. KAS does not: since the harmonic mean is monotone in each argument, fixing $b=1$ gives $\mathrm{KAS}\leq 2a/(1+a)$ and, symmetrically, $\mathrm{KAS}\leq 2b/(1+b)$. Hence $\mathrm{KAS}\geq\tau$ enforces a uniform floor on both branches,
\begin{equation}
\min(a,b)\;\geq\;\tfrac{\tau}{2-\tau},
\end{equation}
so a single strong branch cannot compensate for a weak one.

\section{Details of Experiments}

\subsection{Details of Test Model}
\label{appendix:api details}
In this section, we provide the detailed information about the \noindent\textbf{Model Details.} The evaluated models are summarized in Table~\ref{tab:models}. 

\begin{table}[h]
\centering
\caption{Summary of evaluated LLMs.}
\label{tab:models}
\small
\begin{tabular}{p{0.2\columnwidth}|p{0.75\columnwidth}}  
\toprule[1.5pt]
\textbf{Provider} & \textbf{Models} \\
\midrule
OpenAI & gpt-4o-2024-11-20, gpt-4.1-2025-04-14, gpt-5-2025-08-07, o4-mini-2025-04-16, gpt-5.5 \\
\midrule
Anthropic & claude-sonnet-4-5-20250929 \\
\midrule
Google & gemini-3-flash-preview, gemini-3-pro-preview-11-2025 \\
\midrule
Qwen3 & qwen3-max-preview, qwen3-235b-a22b-instruct-2507, qwen3-235b-a22b-thinking-2507, qwen3.5-397b-a17b \\
\midrule
DeepSeek & deepseek-v3.2, deepseek-r1-250528, deepseek-v4-flash, deepseek-v4-pro \\
\bottomrule[1.5pt]
\end{tabular}
\end{table}

\subsection{Implementation Details.}
\label{appendix:Implementation Details}
\paragraph{Implementation Details.}
We evaluate all models zero-shot with greedy decoding wherever
sampling parameters are user-controllable.
For reasoning models, we set \texttt{reasoning\_effort}=\emph{high}
where supported; otherwise, we use the provider's default
thinking budget.
All API-accessed models share an identical prompt template and a
maximum output length of $16{,}384$ tokens.
We use the default system prompt embedded in each model's chat
template and perform no fine-tuning or test-time adaptation.
For \textit{Pass Rate}, we use gpt-4.1 to judge whether the agent's final
answer after autonomous tool use is consistent with the reference answer.

\subsection{Experimental Prompts}
\label{appendix:experiment}
We provide the complete prompt templates used in our main experiments, including both task execution prompts and function calling instructions. These prompts ensure consistent evaluation across all models.

\begin{tcolorbox}[
    enhanced,
    breakable,
    colback=violet!3,              
    colframe=violet!30!black,      
    colbacktitle=violet!30!black,  
    coltitle=white,
    boxrule=2pt,
    arc=0mm,
    left=10pt,
    right=10pt,
    top=10pt,
    bottom=10pt,
    fonttitle=\bfseries\large,
    title={Prompt Call Agent Prompt},
    attach boxed title to top left={yshift=-2mm}
]

\subsection*{Role}
You are an agent responsible for evaluating if a user's task requires tools.

\subsection*{Task}
\{task\}

\subsection*{Available Tools}
\{tool\_info\}

\subsection*{Instructions}
Assess each tool's necessity according to task specifications.
\begin{itemize}[leftmargin=*, topsep=0pt, itemsep=0pt, parsep=0pt]
    \item Set a tool to \texttt{true} if its functionality is essential for task completion.
    \item Set a tool to \texttt{false} if internal capabilities are sufficient.
\end{itemize}

\subsection*{Output Format}
Return a valid JSON object ONLY in the following format:
\begin{lstlisting}
{
  "tool_decisions": {
    "ToolA": true,
    "ToolB": false,
    ...
  }
}
\end{lstlisting}

\end{tcolorbox}

\begin{tcolorbox}[
    enhanced,
    breakable,
    colback=violet!3,              
    colframe=violet!30!black,      
    colbacktitle=violet!30!black,  
    coltitle=white,                
    boxrule=2pt,
    arc=0mm,
    left=10pt,
    right=10pt,
    top=10pt,
    bottom=10pt,
    fonttitle=\bfseries\large,
    title={Function Call Agent Prompt},
    attach boxed title to top left={yshift=-2mm}
]

\begin{verbatim}
{task}
\end{verbatim}

\end{tcolorbox}

\begin{tcolorbox}[
    enhanced,
    breakable,
    colback=violet!3,
    colframe=violet!40!black,
    colbacktitle=violet!40!black,
    coltitle=white,
    boxrule=1.5pt,
    arc=1mm,
    left=12pt,
    right=12pt,
    top=10pt,
    bottom=10pt,
    fonttitle=\bfseries\large,
    title={Agent Evaluation Prompt},
    attach boxed title to top left={yshift=-2mm, xshift=2mm}
]

\textbf{Role}\par\smallskip
You are an \textbf{Expert Agent Evaluator}.
Your objective is to strictly assess whether an AI agent has successfully fulfilled a user's request based on its execution trajectory and final response.

\medskip
\textbf{Input Data}\par\smallskip
\texttt{\{sub\_questions\_section\}} (User Task)\\
\texttt{\{trajectory\_str\}} (Tool Calling Trajectory)\\
\texttt{\{model\_answer\}} (Agent's Final Answer)\\
\texttt{\{reference\_section\}} (Ground Truth/Reference)

\medskip
\textbf{Task Description}\par\smallskip
Analyze the provided input data to determine if the user's task was solved.
You must generate a reasoning string explaining your judgment and assign a definitive binary status.

\medskip
\textbf{Evaluation Criteria}\par\smallskip
\begin{itemize}
    \item \textbf{Solved}: The agent's final answer is correct \textbf{OR} the answer adequately addresses the user's task.
    \item \textbf{Unsolved}: The agent's final answer is incorrect \textbf{AND} fails to address the user's task, \textbf{OR} the agent refuses the task.
\end{itemize}

\medskip
\textbf{Reasoning Process}\par\smallskip
Your evaluation should follow these logical steps:
\begin{enumerate}
    \item \textbf{Intent Analysis}: Understand the specific goal in the \textit{User Task}.
    \item \textbf{Trajectory Review}: Examine the \textit{Tool Calling Trajectory} to see if the agent took appropriate actions.
    \item \textbf{Answer Verification}: Compare the \textit{Agent's Final Answer} against the \textit{Reference} (if available) and the Solved/Unsolved criteria defined above.
\end{enumerate}

\medskip
\textbf{Output Format}\par\smallskip
You must respond with a single, valid JSON object containing exactly these keys:
\begin{itemize}
    \item \texttt{"content"}: (String) Your reasoning explaining why you assigned the status.
    \item \texttt{"answer\_status"}: (String) Must be exactly "Solved" or "Unsolved".
\end{itemize}

\end{tcolorbox}

\noindent
For the distracting tools experiment, we prioritize randomly selecting distractor tools from the same MCP server as the task-relevant tools to ensure domain and contextual relevance. If insufficient tools are available within that MCP, we supplement by randomly sampling from other MCP servers until forming a complete set of 10 distractor tools. This prioritized sampling strategy guarantees both semantic plausibility of the distractors and sufficient tool diversity, enabling robust evaluation of model tool selection capabilities in realistic multi-tool environments.

\subsection{Additional Experiments}
\label{appendix:Additional Experiments}

To complement the main results, this appendix provides two extended analyses: (i) a fine-grained directional decomposition of knowing/acting errors into tool \textit{Underuse} and \textit{Overuse} (Table~\ref{tab:overall-tool-overuse-underuse}), and (ii) the complete per-hop-type breakdown of KAS underlying the trend reported in RQ3 (Tables~\ref{tab:know-act-kas-external}--\ref{tab:know-act-kas-internal}).

\begin{table*}[htbp]
    \centering
    \resizebox{\textwidth}{!}{%
    \begin{tabular}{l|cccc|cccc|cccc}
        \toprule
        \multirow{3}{*}{\textbf{Model}} & \multicolumn{4}{c|}{\textit{\textbf{External Function}}} & \multicolumn{4}{c|}{\textit{\textbf{Hybrid Composition}}} & \multicolumn{4}{c}{\textit{\textbf{Internal Function}}} \\
        \cmidrule(lr){2-5} \cmidrule(lr){6-9} \cmidrule(lr){10-13}
        & \multicolumn{2}{c}{\textbf{Know}} & \multicolumn{2}{c|}{\textbf{Act}} & \multicolumn{2}{c}{\textbf{Know}} & \multicolumn{2}{c|}{\textbf{Act}} & \multicolumn{2}{c}{\textbf{Know}} & \multicolumn{2}{c}{\textbf{Act}} \\
        \cmidrule(lr){2-3} \cmidrule(lr){4-5} \cmidrule(lr){6-7} \cmidrule(lr){8-9} \cmidrule(lr){10-11} \cmidrule(lr){12-13}
         & \textbf{Underuse} & \textbf{Overuse} & \textbf{Underuse} & \textbf{Overuse} & \textbf{Underuse} & \textbf{Overuse} & \textbf{Underuse} & \textbf{Overuse} & \textbf{Underuse} & \textbf{Overuse} & \textbf{Underuse} & \textbf{Overuse} \\
        \midrule
        
        \multicolumn{13}{c}{\textit{\textbf{Closed-source Models}}} \\
        \midrule
        \rowcolor[gray]{.95} \multicolumn{13}{c}{\textit{Instruct Models}} \\
        gpt-4o                       & 0.0059 & 0.0000 & 0.1613 & 0.0000 & 0.0032 & 0.4574 & 0.3365 & 0.2917 & 0.0000 & 0.7563 & 0.0000 & 0.5352 \\
        gpt-4.1                      & 0.0097 & 0.0000 & 0.1538 & 0.0000 & 0.0055 & 0.4459 & 0.3437 & 0.2858 & 0.0000 & 0.7286 & 0.0000 & 0.4757 \\
        gemini-3-flash               & 0.0065 & 0.0000 & 0.1508 & 0.0332 & 0.0059 & 0.3911 & 0.3421 & 0.2952 & 0.0000 & 0.7211 & 0.0000 & 0.5888 \\
        claude-sonnet-4.5-instruct   & 0.0113 & 0.0000 & 0.0742 & 0.0000 & 0.0055 & 0.4796 & 0.2087 & 0.3483 & 0.0000 & 0.7086 & 0.0000 & 0.7701 \\
        Qwen3-Max-Instruct           & 0.0118 & 0.0000 & 0.0398 & 0.0008 & 0.0041 & 0.4842 & 0.0589 & 0.3867 & 0.0000 & 0.8668 & 0.0000 & 0.8141 \\
        \cdashline{1-13}[0.5pt/2pt]
        \rowcolor[gray]{.95} \multicolumn{13}{c}{\textit{Reasoning Models}} \\
        gpt-5                        & 0.0344 & 0.0000 & 0.1855 & 0.0000 & 0.1033 & 0.0372 & 0.3654 & 0.2681 & 0.0000 & 0.0578 & 0.0000 & 0.4146 \\
                o4-mini                      & 0.0312 & 0.0000 & 0.1731 & 0.0000 & 0.0213 & 0.2853 & 0.3836 & 0.2214 & 0.0000 & 0.3995 & 0.0000 & 0.4507 \\
        gpt-5.5                      & 0.0172 & 0.0032 & 0.0608 & 0.0000 & 0.0063 & 0.1386 & 0.1567 & 0.3428 & 0.0000 & 0.3191 & 0.0000 & 0.6734 \\

        gemini-3-pro                 & 0.0199 & 0.0000 & 0.1376 & 0.0000 & 0.0221 & 0.2365 & 0.1096 & 0.3175 & 0.0000 & 0.6332 & 0.0000 & 0.5875 \\
        claude-sonnet-4.5-think   & 0.0161 & 0.0000 & 0.0366 & 0.0011 & 0.0104 & 0.3837 & 0.2592 & 0.1877 & 0.0000 & 0.5528 & 0.0000 & 0.8556 \\
        \midrule
        
        \multicolumn{13}{c}{\textit{\textbf{Open-source Models}}} \\
        \midrule
        \rowcolor[gray]{.95} \multicolumn{13}{c}{\textit{Instruct Models}} \\
        qwen3-235b-instruct          & 0.0129 & 0.0000 & 0.4274 & 0.0000 & 0.0018 & 0.4638 & 0.2268 & 0.3337 & 0.0000 & 0.9196 & 0.0000 & 0.5452 \\
        qwen3.5-397b-instruct        & 0.0102 & 0.0011 & 0.0538 & 0.0016 & 0.0100 & 0.4144 & 0.1268 & 0.4135 & 0.0000 & 0.6030 & 0.0000 & 0.7789 \\
        deepseek-v3.2                & 0.0129 & 0.0000 & 0.1586 & 0.0000 & 0.0063 & 0.4656 & 0.2042 & 0.3342 & 0.0000 & 0.8593 & 0.0000 & 0.7994 \\
        deepseek-v4-flash            & 0.0172 & 0.0000 & 0.0586 & 0.0032 & 0.0113 & 0.4135 & 0.1132 & 0.4189 & 0.0000 & 0.6683 & 0.0000 & 0.8342 \\
        \cdashline{1-13}[0.5pt/2pt]
        \rowcolor[gray]{.95} \multicolumn{13}{c}{\textit{Reasoning Models}} \\
        qwen3-235b-think             & 0.0262 & 0.0000 & 0.0300 & 0.0000 & 0.0028 & 0.4523 & 0.2781 & 0.2676 & 0.0000 & 0.7538 & 0.0000 & 0.8107 \\
        qwen3.5-397b-think           & 0.0102 & 0.0011 & 0.0414 & 0.0000 & 0.0100 & 0.4131 & 0.0507 & 0.4520 & 0.0000 & 0.5980 & 0.0000 & 0.7764 \\
        deepseek-r1                  & 0.0140 & 0.0000 & 0.4408 & 0.0000 & 0.0127 & 0.3152 & 0.3596 & 0.2803 & 0.0000 & 0.5905 & 0.0000 & 0.6838 \\
        deepseek-v4-pro              & 0.0145 & 0.0000 & 0.0253 & 0.0038 & 0.0027 & 0.4181 & 0.0326 & 0.4339 & 0.0000 & 0.6658 & 0.0000 & 0.8267 \\
        \bottomrule
    \end{tabular}%
    }
    \caption{Overall knowing and acting tool \textit{Underuse} and \textit{Overuse} under three settings.}
    \label{tab:overall-tool-overuse-underuse}
\end{table*}

\paragraph{Directional errors are setting-specific, and reasoning reshapes their distribution.}
Building on the decomposition $1-\mathrm{Acc}_{s}=\mathrm{ToolUnder}_{s}+\mathrm{ToolOver}_{s}$ established in Appendix~\ref{appendix:metrics}, Table~\ref{tab:overall-tool-overuse-underuse} shows that knowing- and acting-side errors are dominated by qualitatively different directions across the three settings.
In \textit{\textbf{External Function}}, errors are almost exclusively \textit{Underuse}: knowing-side overuse is essentially zero, and acting-side mistakes concentrate on omitting required tools (e.g., qwen3-235b-instruct $0.4274$ and deepseek-r1 $0.4408$ Act Underuse), confirming that the bottleneck lies in execution rather than recognition.
In \textit{\textbf{Hybrid Composition}}, all four axes activate simultaneously: a large knowing-side \textit{Overuse} (typically $0.30$--$0.48$ for instruct models) reveals that models systematically over-estimate which subtasks require tools, while acting-side mistakes split roughly evenly between under- and over-invocation, signaling that selective tool use is genuinely two-sided in difficulty.
In \textit{\textbf{Internal Function}}, the picture inverts to a pure \textit{Overuse} regime: knowing-side overuse climbs as high as $0.9196$ (qwen3-235b-instruct) and acting-side overuse as high as $0.8556$ (claude-sonnet-4.5-think). Reasoning closes most of the knowing-side gap (e.g., gpt-5 $\rightarrow 0.0578$) yet leaves the acting side largely intact ($0.4146$ for the same model), corroborating the RQ1 observation that the thinking stage primarily improves \emph{knowing}, while suppressing unnecessary invocations remains an unresolved acting failure.

\begin{table*}[t]
    \centering
    \scriptsize
    \resizebox{\textwidth}{!}{%
        \begin{tabular}{l|*{3}{c}|*{3}{c}|*{3}{c}}
            \toprule
            \multirow{3}{*}{\textbf{Model}} & \multicolumn{9}{c}{\textit{\textbf{External Function}}} \\
            \cmidrule(lr){2-10}
            & \multicolumn{3}{c|}{\textit{single-hop}} & \multicolumn{3}{c|}{\textit{parallel}} & \multicolumn{3}{c}{\textit{multi-hop}} \\
            \cmidrule(lr){2-4} \cmidrule(lr){5-7} \cmidrule(lr){8-10}
            & $\text{Acc}_{\text{know}}$ & $\text{Acc}_{\text{act}}$ & KAS & $\text{Acc}_{\text{know}}$ & $\text{Acc}_{\text{act}}$ & KAS & $\text{Acc}_{\text{know}}$ & $\text{Acc}_{\text{act}}$ & KAS \\
            \midrule

            \multicolumn{10}{c}{\textit{\textbf{Closed-source Models}}} \\
            \midrule
            \rowcolor[gray]{.95} \multicolumn{10}{c}{\textit{Instruct Models}} \\
            gpt-4o & 100.00 & 88.68 & 94.00 & 100.00 & 98.63 & 99.31 & 98.60 & 71.76 & 83.07 \\
            gpt-4.1 & 100.00 & 86.79 & 92.93 & 100.00 & 98.63 & 99.31 & 97.71 & 75.06 & 84.90 \\
            gemini-3-flash & 100.00 & 82.13 & 90.19 & 99.32 & 97.37 & 98.34 & 98.85 & 72.38 & 83.57 \\
            claude-sonnet-4.5-instruct & 100.00 & 97.17 & 98.56 & 99.32 & 98.63 & 98.97 & 97.71 & 85.50 & 91.20 \\
            qwen3-max & 100.00 & 98.11 & 99.05 & 100.00 & 98.63 & 99.31 & 97.20 & 92.68 & 94.89 \\
            \cdashline{1-10}[0.5pt/2pt]
            \rowcolor[gray]{.95} \multicolumn{10}{c}{\textit{Reasoning Models}} \\
            gpt-5 & 100.00 & 83.96 & 91.28 & 97.95 & 94.52 & 96.20 & 93.00 & 72.14 & 81.25 \\
            o4-mini & 99.06 & 88.68 & 93.58 & 95.89 & 97.95 & 96.91 & 95.67 & 69.34 & 80.40 \\
            gpt-5.5 & 100.00 & 98.11 & 99.05 & 96.58 & 97.26 & 96.92 & 97.07 & 88.68 & 92.69 \\
            gemini-3-pro & 100.00 & 93.40 & 96.59 & 97.95 & 98.63 & 98.29 & 96.44 & 73.54 & 83.45 \\
            claude-sonnet-4.5-think & 99.06 & 99.06 & 99.06 & 99.32 & 99.54 & 99.43 & 97.33 & 92.11 & 94.65 \\

            \midrule
            \multicolumn{10}{c}{\textit{\textbf{Open-source Models}}} \\
            \midrule
            \rowcolor[gray]{.95} \multicolumn{10}{c}{\textit{Instruct Models}} \\
            qwen3-235b-instruct & 100.00 & 61.32 & 76.02 & 98.63 & 50.68 & 66.96 & 97.71 & 57.63 & 72.50 \\
            qwen3.5-397b-instruct & 100.00 & 98.58 & 99.28 & 98.86 & 100.00 & 99.43 & 97.96 & 88.04 & 92.74 \\
            deepseek-v3.2 & 100.00 & 85.85 & 92.39 & 100.00 & 86.30 & 92.65 & 96.95 & 81.55 & 88.59 \\
            deepseek-v4-flash & 99.06 & 98.11 & 98.58 & 99.32 & 98.63 & 98.97 & 97.07 & 87.66 & 92.13 \\
            \cdashline{1-10}[0.5pt/2pt]
            \rowcolor[gray]{.95} \multicolumn{10}{c}{\textit{Reasoning Models}} \\
            qwen3-235b-think & 100.00 & 100.00 & 100.00 & 96.88 & 100.00 & 98.42 & 95.54 & 92.90 & 94.20 \\
            qwen3.5-397b-think & 100.00 & 100.00 & 100.00 & 98.86 & 100.00 & 99.43 & 97.96 & 90.20 & 93.92 \\
            deepseek-r1 & 100.00 & 49.06 & 65.83 & 99.32 & 58.90 & 73.95 & 97.07 & 59.80 & 74.01 \\
            deepseek-v4-pro & 100.00 & 99.53 & 99.76 & 98.63 & 100.00 & 99.31 & 97.33 & 93.51 & 95.38 \\

            \bottomrule
        \end{tabular}%
    }
    \caption{$\text{Acc}_{\text{know}}$, $\text{Acc}_{\text{act}}$, and KAS (\%) for \textit{External Function}, broken down by logical type.}
    \label{tab:know-act-kas-external}
\end{table*}

\begin{table*}[t]
    \centering
    \scriptsize
    \resizebox{0.67\textwidth}{!}{%
        \begin{tabular}{l|*{3}{c}|*{3}{c}}
            \toprule
            \multirow{3}{*}{\textbf{Model}} & \multicolumn{6}{c}{\textit{\textbf{Hybrid Composition}}} \\
            \cmidrule(lr){2-7}
            & \multicolumn{3}{c|}{\textit{parallel}} & \multicolumn{3}{c}{\textit{multi-hop}} \\
            \cmidrule(lr){2-4} \cmidrule(lr){5-7}
            & $\text{Acc}_{\text{know}}$ & $\text{Acc}_{\text{act}}$ & KAS & $\text{Acc}_{\text{know}}$ & $\text{Acc}_{\text{act}}$ & KAS \\
            \midrule

            \multicolumn{7}{c}{\textit{\textbf{Closed-source Models}}} \\
            \midrule
            \rowcolor[gray]{.95} \multicolumn{7}{c}{\textit{Instruct Models}} \\
            gpt-4o & 51.46 & 50.29 & 50.87 & 54.39 & 34.78 & 42.43 \\
            gpt-4.1 & 54.97 & 50.29 & 52.53 & 55.04 & 34.62 & 42.50 \\
            gemini-3-flash & 70.18 & 47.08 & 56.35 & 58.52 & 34.30 & 43.25 \\
            claude-sonnet-4.5-instruct & 49.71 & 47.95 & 48.81 & 51.82 & 43.62 & 47.37 \\
            qwen3-max & 49.71 & 50.00 & 49.85 & 51.45 & 56.43 & 53.83 \\
            \cdashline{1-7}[0.5pt/2pt]
            \rowcolor[gray]{.95} \multicolumn{7}{c}{\textit{Reasoning Models}} \\
            gpt-5 & 88.60 & 49.71 & 63.69 & 85.48 & 34.19 & 48.84 \\
            o4-mini & 69.30 & 50.88 & 58.68 & 69.35 & 37.41 & 48.60 \\
            gpt-5.5 & 87.43 & 48.83 & 62.66 & 85.16 & 50.27 & 63.22 \\
            gemini-3-pro & 82.16 & 47.08 & 59.86 & 72.67 & 59.16 & 65.22 \\
            claude-sonnet-4.5-think & 46.88 & 50.00 & 48.39 & 63.11 & 56.27 & 59.49 \\

            \midrule
            \multicolumn{7}{c}{\textit{\textbf{Open-source Models}}} \\
            \midrule
            \rowcolor[gray]{.95} \multicolumn{7}{c}{\textit{Instruct Models}} \\
            qwen3-235b-instruct & 54.68 & 47.66 & 50.93 & 53.22 & 43.27 & 47.73 \\
            qwen3.5-397b-instruct & 52.34 & 48.83 & 50.52 & 58.52 & 45.44 & 51.16 \\
            deepseek-v3.2 & 52.34 & 46.49 & 49.24 & 52.89 & 46.09 & 49.26 \\
            deepseek-v4-flash & 55.85 & 48.54 & 51.94 & 57.82 & 46.46 & 51.52 \\
            \cdashline{1-7}[0.5pt/2pt]
            \rowcolor[gray]{.95} \multicolumn{7}{c}{\textit{Reasoning Models}} \\
            qwen3-235b-think & 54.09 & 41.52 & 46.98 & 54.57 & 46.15 & 50.01 \\
            qwen3.5-397b-think & 53.22 & 47.95 & 50.45 & 58.52 & 50.05 & 53.95 \\
            deepseek-r1 & 68.71 & 43.92 & 53.59 & 66.93 & 34.57 & 45.59 \\
            deepseek-v4-pro & 53.22 & 48.54 & 50.77 & 58.79 & 54.23 & 56.42 \\

            \bottomrule
        \end{tabular}%
    }
    \caption{$\text{Acc}_{\text{know}}$, $\text{Acc}_{\text{act}}$, and KAS (\%) for \textit{Hybrid Composition}, broken down by logical type.}
    \label{tab:know-act-kas-hybrid}
\end{table*}

\begin{table*}[t]
    \centering
    \scriptsize
    \resizebox{\textwidth}{!}{%
        \begin{tabular}{l|*{3}{c}|*{3}{c}|*{3}{c}}
            \toprule
            \multirow{3}{*}{\textbf{Model}} & \multicolumn{9}{c}{\textit{\textbf{Internal Function}}} \\
            \cmidrule(lr){2-10}
            & \multicolumn{3}{c|}{\textit{single-hop}} & \multicolumn{3}{c|}{\textit{parallel}} & \multicolumn{3}{c}{\textit{multi-hop}} \\
            \cmidrule(lr){2-4} \cmidrule(lr){5-7} \cmidrule(lr){8-10}
            & $\text{Acc}_{\text{know}}$ & $\text{Acc}_{\text{act}}$ & KAS & $\text{Acc}_{\text{know}}$ & $\text{Acc}_{\text{act}}$ & KAS & $\text{Acc}_{\text{know}}$ & $\text{Acc}_{\text{act}}$ & KAS \\
            \midrule

            \multicolumn{10}{c}{\textit{\textbf{Closed-source Models}}} \\
            \midrule
            \rowcolor[gray]{.95} \multicolumn{10}{c}{\textit{Instruct Models}} \\
            gpt-4o & 65.94 & 53.62 & 59.15 & 0.00 & 20.77 & 0.00 & 4.62 & 64.62 & 8.62 \\
            gpt-4.1 & 72.46 & 65.22 & 68.65 & 4.62 & 26.15 & 7.85 & 2.31 & 65.13 & 4.46 \\
            gemini-3-flash & 68.12 & 38.41 & 49.12 & 6.15 & 20.00 & 9.41 & 6.92 & 65.13 & 12.51 \\
            claude-sonnet-4.5-instruct & 75.36 & 17.39 & 28.26 & 8.46 & 19.23 & 11.75 & 0.77 & 32.69 & 1.50 \\
            qwen3-max & 38.41 & 21.74 & 27.77 & 0.00 & 3.85 & 0.00 & 0.00 & 30.00 & 0.00 \\
            \cdashline{1-10}[0.5pt/2pt]
            \rowcolor[gray]{.95} \multicolumn{10}{c}{\textit{Reasoning Models}} \\
            gpt-5 & 100.00 & 71.01 & 83.05 & 98.46 & 30.00 & 45.99 & 83.85 & 73.85 & 78.53 \\
            o4-mini & 100.00 & 71.74 & 83.54 & 36.92 & 29.23 & 32.63 & 40.77 & 62.95 & 49.49 \\
            gpt-5.5 & 98.55 & 57.97 & 73.00 & 66.15 & 19.23 & 29.80 & 37.69 & 19.23 & 25.47 \\
            gemini-3-pro & 68.12 & 39.86 & 50.29 & 31.54 & 19.23 & 23.89 & 8.46 & 64.74 & 14.96 \\
            claude-sonnet-4.5-think & 94.20 & 3.62 & 6.97 & 20.00 & 11.92 & 14.94 & 16.92 & 28.46 & 21.22 \\

            \midrule
            \multicolumn{10}{c}{\textit{\textbf{Open-source Models}}} \\
            \midrule
            \rowcolor[gray]{.95} \multicolumn{10}{c}{\textit{Instruct Models}} \\
            qwen3-235b-instruct & 23.19 & 46.38 & 30.92 & 0.00 & 43.08 & 0.00 & 0.00 & 46.92 & 0.00 \\
            qwen3.5-397b-instruct & 86.96 & 31.16 & 45.88 & 9.23 & 16.15 & 11.75 & 19.23 & 17.69 & 18.43 \\
            deepseek-v3.2 & 36.96 & 5.80 & 10.03 & 0.00 & 19.23 & 0.00 & 3.85 & 36.03 & 6.96 \\
            deepseek-v4-flash & 81.16 & 15.22 & 25.63 & 11.54 & 16.15 & 13.46 & 3.85 & 17.69 & 6.32 \\
            \cdashline{1-10}[0.5pt/2pt]
            \rowcolor[gray]{.95} \multicolumn{10}{c}{\textit{Reasoning Models}} \\
            qwen3-235b-think & 60.14 & 13.04 & 21.43 & 6.15 & 15.38 & 8.79 & 5.38 & 28.72 & 9.06 \\
            qwen3.5-397b-think & 86.23 & 31.16 & 45.78 & 10.00 & 16.92 & 12.57 & 20.77 & 18.46 & 19.55 \\
            deepseek-r1 & 76.81 & 23.91 & 36.47 & 27.69 & 35.00 & 30.92 & 16.15 & 36.41 & 22.38 \\
            deepseek-v4-pro & 77.54 & 18.84 & 30.31 & 9.23 & 16.15 & 11.75 & 9.23 & 16.92 & 11.94 \\

            \bottomrule
        \end{tabular}%
    }
    \caption{$\text{Acc}_{\text{know}}$, $\text{Acc}_{\text{act}}$, and KAS (\%) for \textit{Internal Function}, broken down by logical type}
    \label{tab:know-act-kas-internal}
\end{table*}

\paragraph{KAS degrades with logical complexity, with setting-dependent slopes.}
Tables~\ref{tab:know-act-kas-external}--\ref{tab:know-act-kas-internal} report the full per-hop-type breakdown underlying Figure~\ref{fig:logic_types}.
In \textit{\textbf{External Function}}, the decline is moderate and graceful: KAS slides from $\sim$95 on single-hop to $\sim$87 on multi-hop, and closed-source reasoning models (e.g., claude-sonnet-4.5-think $99.06/99.43/94.65$ on single/parallel/multi-hop) stay above $90$ throughout, indicating that execution-side reasoning scales well when the capability boundary is unambiguous.
In \textit{\textbf{Hybrid Composition}}, both parallel and multi-hop lie in the $45$--$65$ range, and the reasoning advantage widens with depth: gpt-5.5 reaches $63.22$ on multi-hop versus $42.50$ for gpt-4.1, suggesting that explicit reasoning is most valuable when the model must \emph{localize} the boundary across multiple sub-decisions rather than apply it once.
In \textit{\textbf{Internal Function}}, the degradation becomes catastrophic: nearly all instruct models collapse to near-zero KAS on the parallel and multi-hop subsets (qwen3-max $0.00/0.00$, gpt-4o $0.00/8.62$), and only the strongest reasoning models retain non-trivial competence (gpt-5 $83.05/45.99/78.53$). Crucially, weaker models collapse uniformly on both parallel and multi-hop, suggesting that any compositional structure triggers the same overuse failure; only the strongest reasoning models begin to differentiate the two (gpt-5 $45.99$ vs.\ $78.53$), implying that simultaneous-boundary judgments and chained reasoning constitute distinct difficulties that surface only after the basic restraint floor is achieved.

\end{document}